\documentclass[twoside,11pt]{article}

\usepackage{blindtext}

\usepackage{rotating}
\usepackage{pdflscape}
\usepackage{booktabs}
\usepackage{multirow}
\usepackage{paralist,amsmath, amssymb,bm}
\usepackage{algorithm,algorithmic}
\usepackage[preprint]{jmlr2e}

\makeatletter
\AtBeginDocument{\Hy@breaklinkstrue}
\makeatother

\newtheorem{thm}{Theorem}
\newtheorem{prop}{Proposition}
\newtheorem{lem}{Lemma}
\newtheorem{cor}{Corollary}
\newtheorem{ass}{Assumption}

\def \y {\mathbf{y}}

\def \x {\mathbf{x}}
\def \g {\mathbf{g}}

\def \u {\mathbf{u}}
\def \H {\mathcal{H}}
\def \w {\mathbf{w}}
\def \R {\mathbb{R}}

\def \tg {\tilde{g}}

\def \q {\mathbf{q}}
\def \bv {\mathbf{v}}

\def \p {\mathbf{p}}
\def \q {\mathbf{q}}

\def \dtb  {\bar\Delta}

\def \xb {\bar{\x}}

\def \Ot {\widetilde{O}}

\def \Rt {\mathcal{R}}

\def \X {\mathcal{X}}

\newcommand{\tabincell}[2]{\begin{tabular}{@{}#1@{}}#2\end{tabular}}

\DeclareMathOperator*{\DReg}{D-Regret}

\DeclareMathOperator*{\argmin}{argmin}

\DeclareMathOperator*{\erf}{erf}

\usepackage{color}
\usepackage{lastpage}
\jmlrheading{23}{2022}{1-\pageref{LastPage}}{1/21; Revised 5/22}{9/22}{21-0000}{Lijun Zhang, Linzhe He, Wei Jiang, and Shiyin Lu}

\ShortHeadings{Revisiting Smoothed Online Learning}{Zhang, He, Jiang, and Lu}
\firstpageno{1}

\begin{document}

\title{Revisiting Smoothed Online Learning:\\ Competitive Ratio, Dynamic Regret with Switching Cost, and Best of Both Worlds}

\author{\name Lijun Zhang \email zhanglj@lamda.nju.edu.cn\\
        \name Linzhe He \email helz@lamda.nju.edu.cn\\
       \addr State Key Laboratory of Novel Software Technology, Nanjing University, Nanjing 210023, China
       \addr School of Artificial Intelligence, Nanjing University, Nanjing 210023, China       \AND
        \name Wei Jiang \email 12025220@njust.edu.cn\\
        \addr School of Computer Science and Engineering, Nanjing University of Science and Technology, Nanjing 210094, China
        \AND
        \name Shiyin Lu \email running.lsy@alibaba-inc.com\\
        \addr Alibaba Group, Hangzhou 310052, China
}

\editor{My editor}
\maketitle
\begin{abstract}
We revisit smoothed online learning and develop algorithms with improved competitive ratios, optimal dynamic regret
with switching cost, and best-of-both-worlds guarantees. Surprisingly, minimizing the hitting cost alone is
$\max(1,\frac{2}{\alpha})$-competitive for $\alpha$-polyhedral functions under the $\ell_2$-norm
switching cost and $1+\frac{4}{\lambda}$-competitive for $\lambda$-quadratic growth functions under the squared
$\ell_2$-norm switching cost, significantly improving existing guarantees.
The polyhedral bound is tight for this naive algorithm, and we establish a matching lower bound for $\alpha\geq1$.
For convex $\lambda$-quadratic growth hitting costs, minimizing a suitably weighted sum of hitting and squared
$\ell_2$-norm switching costs yields a competitive ratio of $1+\frac{2}{\sqrt{\lambda}}$.
For dynamic regret with the $\ell_2$-norm switching cost, we develop Smoothed Ader (SAder), which achieves the
optimal $O(\sqrt{T(1+P_T)})$ rate in the standard online convex optimization setting, where $P_T$ is the
comparator's path-length. Its lookahead version removes the bounded-gradient assumption, and a matching lower bound establishes
optimality in this setting. Finally, for convex hitting costs on a bounded convex domain, we develop UDNP to aggregate any two experts in the lookahead setting under both switching costs. Its additional cumulative cost is at most $O(1)$
relative to the first expert and $O(\sqrt{T\log T})$ relative to the second, yielding best-of-both-worlds guarantees without bounded gradients.
\end{abstract}


\section{Introduction}
Online learning is the process of making a sequence of predictions given knowledge of the answers to previous tasks and possibly additional  information \citep{Online:suvery}. While traditional online learning aims to make the prediction as accurate as possible, in this paper, we study smoothed online learning (SOL), where the online learner incurs a switching cost for changing its predictions between rounds \citep{NIPS2013_5151}. SOL has received considerable attention recently because in many real-world applications, a change of action usually brings some additional cost. Examples include the dynamic right-sizing for data centers \citep{Dynamic:Data:Center}, geographical load balancing \citep{Geographical:Load:Balance}, real-time electricity pricing \citep{electricity:pricing},  video streaming \citep{video:transport}, spatiotemporal sequence prediction \citep{Spatiotemporal:Prediction}, multi-timescale control \citep{Optimization:timescales}, and thermal management \citep{Management:MPSoCs}.

Specifically, SOL is performed in a sequence of consecutive rounds, where at round $t$ the learner is asked to select a point $\x_t$ from the decision set $\X$, and suffers a hitting cost $f_t(\x_t)$. Depending on the performance metric, the learner \emph{may} be allowed to observe $f_t(\cdot)$ when making decisions, which is different from the traditional online learning in which $f_t(\cdot)$ is revealed only after the learner submits its decision \citep{bianchi-2006-prediction}. Additionally, the learner also incurs a switching cost $m(\x_t,\x_{t-1})$ for changing decisions between successive rounds. Common choices for the switching cost include the $\ell_2$-norm distance $\|\x_t-\x_{t-1}\|$ and the squared $\ell_2$-norm $\|\x_t-\x_{t-1}\|^2/2$ \citep{NIPS2019_8463}. In the literature, there are two performance metrics for SOL: competitive ratio and dynamic regret with switching cost.

Competitive ratio is popular in the community of online algorithms
\citep{Online:Competitive:book}. It is defined as the worst-case ratio
of the total cost incurred by the online learner to the offline
optimal cost:
\begin{equation}\label{eqn:competitive:ratio}
\frac{
    \sum_{t=1}^T \big(f_t(\x_t)+m(\x_t,\x_{t-1})\big)
}{
    \min_{\u_0,\u_1,\ldots,\u_T\in\X}
    \sum_{t=1}^T \big(f_t(\u_t)+m(\u_t,\u_{t-1})\big)
}.
\end{equation}
When focusing on the competitive ratio, we consider the \emph{lookahead} setting, in which the learner observes $f_t(\cdot)$
before picking $\x_t$. The problem is still nontrivial due to the coupling created by the switching cost.

Dynamic regret with switching cost extends dynamic regret
\citep{zinkevich-2003-online} by accounting for switching costs.
Whereas competitive ratio measures a multiplicative comparison
with the offline optimum, dynamic regret measures the additive
difference between the learner's total cost and that of an arbitrary
comparator sequence $\u_0,\u_1,\ldots,\u_T\in\X$:
\begin{equation}\label{eqn:dynamic:regret:switching}
\sum_{t=1}^T \big(f_t(\x_t)+m(\x_t,\x_{t-1})\big)
-
\sum_{t=1}^T \big(f_t(\u_t)+m(\u_t,\u_{t-1})\big).
\end{equation}
We keep the comparator sequence explicit in
\eqref{eqn:dynamic:regret:switching}, rather than minimizing over it
as in previous work \citep{SOCO:OBD,NIPS2019_8463}, and seek bounds
that adapt to its regularity, measured by the path-length
\begin{equation}\label{eqn:path}
P_T=\sum_{t=1}^T \|\u_t-\u_{t-1}\|.
\end{equation}
For dynamic regret with switching cost, we consider both the
standard OCO setting and the lookahead setting.

\paragraph{Motivation}
Online balanced descent (OBD) \citep{SOCO:OBD}
balances the hitting and switching costs by projecting the
previous decision onto a suitably chosen sublevel set of the
current hitting-cost function. OBD and its variants \citep{NIPS2019_8463} exploit
structural properties of the hitting costs to obtain
dimension-free competitive ratios. These results motivate
examining the simpler \emph{greedy} algorithm, which selects
its decision by solving
\begin{equation}\label{eqn:greedy}
    \min_{\x\in\X}\quad
    f_t(\x)+\gamma m(\x,\x_{t-1}),
\end{equation}
where $\gamma\geq0$ is a trade-off parameter.
As a special case of Regularized OBD (R-OBD), a suitably tuned
greedy algorithm achieves the optimal competitive ratio for
strongly convex hitting costs under the squared $\ell_2$-norm
switching cost \citep{NIPS2019_8463}.
Moreover, \citet{10.1145/3379484} analyze the choice $\gamma=0$,
referred to as the \emph{naive} approach, for polyhedral and
quadratic growth functions.
We revisit these simple algorithms to sharpen their competitive
guarantees under different growth conditions and to clarify
when explicitly accounting for switching costs improves performance.

For dynamic regret, neither the guarantee of OBD nor that of
R-OBD provides an adaptive bound for the $\ell_2$-norm switching
cost. OBD requires a prescribed upper bound on the comparator's
path-length, and its regret bound depends on this upper bound
rather than the actual path-length
\citep[Corollary~11]{SOCO:OBD}.
R-OBD provides an adaptive guarantee, but  for the squared
$\ell_2$-norm switching cost \citep{NIPS2019_8463}.
Moreover, both guarantees require lookahead access to the
current hitting cost.
We therefore seek adaptive dynamic-regret guarantees for the
$\ell_2$-norm switching cost under both standard and lookahead
feedback. We also investigate whether a single algorithm can
achieve both a competitive-ratio guarantee and an adaptive
dynamic-regret guarantee, connecting to earlier work on
combining the two criteria
\citep{pmlr-v30-Andrew13,pmlr-v98-daniely19a}.
Expert aggregation offers a natural approach, but directly
applying this approach requires a known bound on the experts'
one-round cost differences. We investigate how lookahead
information can overcome this obstacle without imposing
an additional bounded-gradient assumption.

Our contributions are summarized as follows.
\paragraph{Competitive Ratio} We sharpen the competitive guarantees of the naive and
greedy algorithms and establish complementary lower bounds. For $\alpha$-polyhedral hitting costs and the $\ell_2$-norm
switching cost, the naive algorithm achieves a competitive
ratio of $\max\left(1,\frac{2}{\alpha}\right)$.
This improves upon the $3+\frac{8}{\alpha}$ ratio of OBD
\citep{SOCO:OBD} and the $1+\frac{2}{\alpha}$ bound of
\citet[Lemma~1]{10.1145/3379484}.
We show that this bound is tight for the naive algorithm
and cannot be improved by any fixed choice of $\gamma$
in the greedy algorithm.
Moreover, we establish a matching lower bound for
$\alpha\geq1$.

For $\lambda$-quadratic growth hitting costs and the squared
$\ell_2$-norm switching cost, the naive algorithm achieves
a competitive ratio of $1+\frac{4}{\lambda}$, improving upon
the $\max\left(1+\frac{6}{\lambda},4\right)$ bound of
\citet[Lemma~1]{10.1145/3379484} and matching, as $\lambda$
tends to zero, the order of the lower bound for this algorithm
in \citet[Theorem~5]{NIPS2019_8463}.
Neither this guarantee nor the polyhedral guarantee requires
convexity.
Under additional convexity, a suitable choice of $\gamma>0$
yields a competitive ratio of $1+\frac{2}{\sqrt{\lambda}}$,
improving the naive guarantee for small $\lambda$.
Compared with Greedy OBD \citep[Theorem~3]{NIPS2019_8463},
our guarantee has the same asymptotic order with much smaller
constants, although it requires convexity rather than
quasiconvexity.
Our analysis of the naive and greedy algorithms is also
simpler, avoiding the geometric arguments used for OBD
and Greedy OBD.

\paragraph{Dynamic Regret with Switching Cost} For convex hitting costs and the $\ell_2$-norm switching cost, we develop Smoothed Ader (SAder) by extending Ader \citep{Adaptive:Dynamic:Regret:NIPS}, an algorithm that achieves optimal dynamic regret in standard online convex optimization. Ader uses Hedge \citep{FREUND1997119} to aggregate online gradient descent (OGD) experts \citep{zinkevich-2003-online} with different step sizes. SAder modifies this approach by incorporating the experts' switching costs into the meta-algorithm's losses. Under the standard bounded-domain and bounded-gradient assumptions, SAder achieves an optimal $O(\sqrt{T(1+P_T)})$ bound on the sum of dynamic regret and the learner's switching cost, matching Ader's rate despite the additional switching cost. Compared with OBD \citep[Corollary~11]{SOCO:OBD}, SAder adapts to the actual path-length $P_T$ without requiring a prescribed upper bound on $P_T$ or lookahead feedback.

We further develop lookahead SAder, whose greedy experts
generate predictions using the current hitting cost.
The meta-algorithm updates the aggregation weights using
the experts' current one-round costs before making its decision.
Lookahead SAder achieves the same $O(\sqrt{T(1+P_T)})$
regret bound without assuming bounded gradients.
We establish a matching $\Omega(\sqrt{T(1+P_T)})$ lower
bound in the lookahead setting, showing that this rate
is optimal up to constant factors.

\paragraph{Best of Both Worlds} We develop UDNP, an unbounded-input variant of the Discounted-Normal-Predictor \citep{NIPS2011_11b921ef}, to aggregate any two experts in the lookahead setting. Building on expert aggregation with switching costs \citep{pmlr-v98-daniely19a}, UDNP exploits lookahead information to handle potentially unbounded differences between the experts' one-round costs, without requiring a prescribed bound on these differences or a bounded-gradient assumption. Our approach applies to both the $\ell_2$-norm and squared $\ell_2$-norm switching costs. For convex hitting costs on a bounded convex domain, the learner's total cost exceeds that of the first expert by at most a constant and that of the second expert by at most $O(\sqrt{T\log T})$. Thus, when the first expert provides a competitive-ratio guarantee and the second a dynamic-regret guarantee, UDNP preserves the former up to an additive constant while adding only an $O(\sqrt{T\log T})$ term to the latter.

As concrete applications, we instantiate this approach with experts analyzed in this paper. For convex hitting costs satisfying the $\alpha$-polyhedral condition, combining the naive algorithm with lookahead SAder under the $\ell_2$-norm switching cost yields the competitive factor $\max\left(1,\frac{2}{\alpha}\right)$ up to an additive constant and an $O(\sqrt{T(1+P_T)}+\sqrt{T\log T})$ dynamic-regret bound. For convex hitting costs satisfying the $\lambda$-quadratic growth condition, combining two greedy experts under the squared $\ell_2$-norm switching cost yields the competitive factor $1+\frac{2}{\sqrt{\lambda}}$ up to an additive constant and an $O(P_T+\sqrt{T\log T})$ dynamic-regret bound.

\paragraph{Extensions over the Conference Version} A preliminary version of this paper was presented at the 35th Annual Conference on Neural Information Processing Systems (NeurIPS) \citep{Smoothed:Online:NeurIPS}. The present paper extends that version in the following aspects:
\begin{compactitem}
\item We extend the competitive-ratio analysis for
polyhedral hitting costs. We prove that the $\max\left(1,\frac{2}{\alpha}\right)$ bound
from the conference version is tight for the naive algorithm
and cannot be improved by any fixed choice of $\gamma$ in the greedy algorithm.
We further establish a matching lower bound for
$\alpha\geq1$.

\item We develop UDNP to address the best-of-both-worlds
question left open in the conference version.
For the $\ell_2$-norm switching cost, we establish
general guarantees for aggregating arbitrary pairs
of experts while handling potentially unbounded cost
differences without assuming bounded gradients.
We instantiate these guarantees by combining the naive
algorithm with lookahead SAder.

\item We extend UDNP to the squared $\ell_2$-norm
switching cost through a new interpolation inequality.
The resulting aggregation guarantee also applies to
arbitrary pairs of experts, and we instantiate it with
two greedy experts.
We further establish a dynamic-regret bound for the
greedy algorithm under general convexity, allowing
nonsmooth hitting costs and constrained minimizers.

\item We revise lookahead SAder to update its aggregation
weights using the experts' current one-round costs,
removing the circular dependence on the current
meta-decision. We also provide a proof based on the strong convexity of the experts' regularized objectives, without requiring
differentiability of the hitting costs.
\end{compactitem}

\paragraph{Organization} We review related work in Section~\ref{sec:related} and study the competitive ratios of the naive and greedy algorithms in Section~\ref{sec:ratio}. Section~\ref{sec:regret} develops SAder and its lookahead variant for dynamic regret with switching cost. The best-of-both-worlds framework, UDNP, is presented in Section~\ref{sec:bow} for both the $\ell_2$-norm and squared $\ell_2$-norm switching costs. Proofs of the main results appear in Section~\ref{sec:proof}, with proofs of supporting lemmas deferred to the appendices. We conclude the paper and discuss future directions in Section~\ref{sec:conclusion}.

\section{Related Work}  \label{sec:related}
This section reviews related work on smoothed online learning (SOL), dynamic regret, and online optimization with switching constraints.

\subsection{Smoothed Online Learning}
SOL has been investigated under the setting of multi-armed bandits \citep{SSR:Bandits:Switching,Bandit:Metric:Switching,Dekel:Bandits:Switching,NIPS2017_7000, pmlr-v65-koren17a}, prediction with expert advice \citep{NIPS2013_5151}, and online convex optimization \citep{Dynamic:Data:Center,bansal_et_al:LIPIcs,10.1145/3375788}. In the following, we mainly discuss smoothed online convex optimization (SOCO).

The early works on SOCO are motivated by real-world applications with switching costs, and focus on designing competitive algorithms in the low-dimensional setting \citep{Dynamic:Data:Center}. In particular, \citet{bansal_et_al:LIPIcs} show that for SOCO on the real line, the competitive ratio can be upper bounded by $2$, which is proved to be optimal \citep{10.1007/978-3-319-89441-6_13}. They also establish a competitive ratio of $3$ under the memoryless setting. In the study of SOCO, it is common to assume that the learner has access to predictions of future hitting costs, and several algorithms \citep{Geographical:Load:Balance,10.1145/2745844.2745854,10.1145/2896377.2901464,Prediction:Switching,NEURIPS2020_a6e4f250} have been developed based on receding horizon control (RHC) \citep{RHC:2005}. In fact, the greedy algorithm in (\ref{eqn:greedy}) can be treated as a variant of RHC. However, previous results for RHC are limited to special problems,  and (\ref{eqn:greedy}) remains under-explored.  For example, with a prediction window of length $W$, \citet{Prediction:Switching} derive a dynamic-regret bound with switching cost that decreases exponentially with $W$, under assumptions including strong convexity and smoothness.

One milestone is the online balanced descent (OBD) \citep{SOCO:OBD}, which has a dimension-free competitive ratio even when the learner can only observe the hitting cost of the current round. Specifically, OBD iteratively projects the previous point onto a carefully chosen level set of the hitting cost  so as to balance the switching cost and the hitting cost. When the hitting cost is $\alpha$-polyhedral and convex, and the switching cost is the $\ell_2$-norm distance, OBD attains a $3+\frac{8}{\alpha}$ competitive ratio. Furthermore, OBD can also be tuned to control the dynamic regret with switching cost. Let $L$ be an upper bound of the path-length of the comparator sequence, i.e., $P_T \leq L$. \citet[Corollary 11]{SOCO:OBD} have proved that
\begin{equation} \label{eqn:obd:regret}
\sum_{t=1}^T \big( f_t(\x_t) + \|\x_t-\x_{t-1}\|\big) - \min_{P_T \leq L} \sum_{t=1}^T  \big(f_t (\u_t) + \|\u_t- \u_{t-1}\|\big) = O\left(\sqrt{TL}\right)
\end{equation}
leading to sublinear regret when $L=o(T)$. However, there are two limitations: (i) the value of $L$ needs to be given as a parameter; and  (ii) it is nonadaptive because the upper bound depends on $L$ instead of the actual path-length $P_T$.

\begin{table}[t]
\centering
\caption{Competitive Ratios of OBD and Related Methods. All algorithms operate in the lookahead setting. The switching cost is the $\ell_2$-norm for the $\alpha$-polyhedral results and Constrained OBD, and the squared $\ell_2$-norm for all remaining results.}
\label{tab:results}
\renewcommand{\arraystretch}{1.2}
\begin{tabular}{
@{}c@{\hspace{.5ex}}|
@{\hspace{.5ex}}c@{\hspace{.5ex}}|
@{\hspace{.5ex}}c@{}
}
\toprule
Algorithm & Assumptions & Competitive Ratio \\ \hline
\multirow{3}{*}{\tabincell{c}{OBD\\ \citep{SOCO:OBD} }} & $f_t(\cdot)$ is $\alpha$-polyhedral & \multirow{3}{*}{$\displaystyle 3+\frac{8}{\alpha}$} \\
& $f_t(\cdot)$ is convex & \\ & $\X$ is convex & \\ \hline
The Naive Approach & \multirow{4}{*}{$f_t(\cdot)$ is $\alpha$-polyhedral} & \multirow{2}{*}{$\displaystyle 1+\frac{2}{\alpha}$} \\
\citep[Lemma~1]{10.1145/3379484} & & \\ \cline{1-1} \cline{3-3}
The Naive Approach & &
\multirow{2}{*}{$\displaystyle
\max\left(1,\frac{2}{\alpha}\right)$}
\\
(Theorem~\ref{thm:ratio:poly}) & & \\ \hline

\multirow{3}{*}{\tabincell{c}{Greedy OBD\\\citep[Theorem~3]{NIPS2019_8463}}}
&
$f_t(\cdot)$ is $\lambda$-quadratic growth
&
\multirow{3}{*}{$\displaystyle
O\left(\frac{1}{\sqrt{\lambda}}\right)
\text{ as }\lambda\to0$}
\\
&
$f_t(\cdot)$ is quasiconvex
&\\ & $\X$ is convex
& \\ \hline

The Naive Approach
&
\multirow{4}{*}{$f_t(\cdot)$ is $\lambda$-quadratic growth}
&
\multirow{2}{*}{$\displaystyle
\max\left(1+\frac{6}{\lambda},4\right)$}
\\
\citep[Lemma~1]{10.1145/3379484}
&
&
\\ \cline{1-1} \cline{3-3}

The Naive Approach
&
&
\multirow{2}{*}{$\displaystyle 1+\frac{4}{\lambda}$}
\\
(Theorem~\ref{thm:naive:quadratic})
&
&
\\ \hline

\multirow{3}{*}{
\tabincell{c}{
The Greedy Algorithm\\
(Theorem~\ref{thm:greedy:convex:quadratic})
}
}
&
$f_t(\cdot)$ is $\lambda$-quadratic growth
&
\multirow{3}{*}{$\displaystyle
1+\frac{2}{\sqrt{\lambda}}$}
\\
&
$f_t(\cdot)$ is convex
&
\\
&
$\X$ is convex
&
\\ \hline

OBD
&
\multirow{4}{*}{
\tabincell{c}{
$f_t(\cdot)$ is $\lambda$-strongly convex\\
$\X$ is convex
}
}
&
\multirow{2}{*}{$\displaystyle
3+O\left(\frac{1}{\lambda}\right)$}
\\
\citep{OBD:Strongly}
&
&
\\ \cline{1-1} \cline{3-3}

The Greedy Algorithm and R-OBD
&
&
\multirow{2}{*}{$\displaystyle
\frac12+\frac12\sqrt{1+\frac{4}{\lambda}}$}
\\
\citep[Theorem~7]{NIPS2019_8463}
&
&
\\ \hline

\multirow{4}{*}{
\tabincell{c}{
Constrained OBD\\
\citep{pmlr-v125-argue20a}}} & $f_t(\cdot)$ is $\lambda$-strongly convex &
\multirow{4}{*}{$\displaystyle
4+4\sqrt{\frac{2H}{\lambda}}$}
\\
&
$f_t(\cdot)$ is $H$-smooth
&
\\
&
$\min_{\x\in\R^d}f_t(\x)=0$
&
\\
&
$\X$ is convex
&
\\ \bottomrule
\end{tabular}

\end{table}

Later, \citet{OBD:Strongly} demonstrate that OBD is $3+O(\frac{1}{\lambda})$-competitive for $\lambda$-strongly convex functions, when the switching cost is set to be the squared $\ell_2$-norm. In a subsequent work, \citet[Theorems~4 and~7]{NIPS2019_8463} propose Regularized OBD (R-OBD), which improves the competitive ratio to $\frac{1}{2}+ \frac{1}{2}\sqrt{1+\frac{4}{\lambda}}$, matching the lower bound of strongly convex functions exactly  \cite[Theorem 1]{NIPS2019_8463}. R-OBD includes  (\ref{eqn:greedy}) as a special case, which also enjoys the optimal competitive ratio for strongly convex functions. Furthermore, \citet[Theorem~6]{NIPS2019_8463}
analyze the dynamic regret of R-OBD, and the following
bound can be extracted from their proof by setting
$\lambda_1=1$ and $\lambda_2=0$:
\[
\sum_{t=1}^T \left( f_t(\x_t) + \frac{1}{2}\|\x_t-\x_{t-1}\|^2 \right) -  \sum_{t=1}^T  \left(f_t (\u_t) + \frac{1}{2}\|\u_t- \u_{t-1}\|^2\right) = O(1+P_T ).
\]
Compared with \eqref{eqn:obd:regret}, this bound is adaptive because it depends on the actual path-length of the
comparator sequence. However, this squared-norm guarantee does not cover the $\ell_2$-norm switching cost.
Moreover, the proof uses a vanishing gradient condition for the regularized objective, which need not hold at
a constrained minimizer.  When the hitting cost is both quasiconvex and $\lambda$-quadratic growth, \citet[Theorem 3]{NIPS2019_8463} have demonstrated that their Greedy OBD algorithm attains an $O(1/\sqrt{\lambda})$ competitive ratio, as $\lambda \rightarrow  0$.

\citet{10.1145/3379484} have analyzed the naive approach which ignores the switching cost and simply minimizes the hitting cost in each round, i.e., the greedy algorithm with $\gamma=0$. It is a bit surprising that this naive approach is $1+\frac{2}{\alpha}$-competitive for $\alpha$-polyhedral functions and $\max(1+\frac{6}{\lambda},4)$-competitive for $\lambda$-quadratic growth functions, without any convexity assumption \citep[Lemma 1]{10.1145/3379484}. \citet{pmlr-v125-argue20a} have investigated a hybrid setting in which the hitting cost is both $\lambda$-strongly convex and $H$-smooth, but the switching cost is the $\ell_2$-norm instead of the squared one. They develop Constrained OBD, and establish a $4+ 4\sqrt{2H/\lambda}$ competitive ratio. However, their analysis requires the hitting costs to be globally nonnegative, rather than merely nonnegative on the feasible set $\X$.  To facilitate comparisons, we summarize the competitive ratios of OBD, its variants, and the greedy algorithm in Table~\ref{tab:results}.

SOCO is closely related to convex body chasing (CBC), where the
learner selects a point from each arriving convex body and seeks
to minimize the total movement
\citep{Convex:Body,CCBF:2016,NCB:2018,Linear:CNCB,CCCB:Bubeck,Optimal:CNCB}.
For general CBC in Euclidean space, algorithms achieve a
competitive ratio of $O(\min\{d,\sqrt{d\log T}\})$
\citep{Linear:CCB,Optimal:CCB}, while a lower bound of $\Omega(\sqrt{d})$ is known \citep{Convex:Body}. Recent work investigates how to incorporate advice into competitive algorithms for SOCO and related chasing problems while retaining worst-case guarantees. \citet{pmlr-v178-christianson22a} combine black-box advice with
a competitive algorithm for convex body and function chasing;
for polyhedral hitting costs, they use the naive approach from
our conference version \citep{Smoothed:Online:NeurIPS} as the
competitive baseline.
For possibly nonconvex $\alpha$-polyhedral hitting costs,
\citet{Rutten2023Unreliable} establish near-optimal tradeoffs
between consistency and robustness with untrusted predictions.
\citet{pmlr-v206-christianson23a} further characterize such
tradeoffs in the more general framework of metrical task systems.

Recent work also develops dynamic-regret guarantees for SOCO
in several directions.
\citet{Smooth:DNP:NeurIPS} obtain interval-wise regret guarantees
using Discounted-Normal-Predictor \citep{NIPS2011_11b921ef},
while \citet{Yang2025DelayedSOCO} consider delayed gradient feedback.
\citet{pmlr-v300-mhaisen26a} introduce partially lazy gradient
updates and aggregate the resulting experts using the SAder
meta-algorithm from our conference version
\citep{Smoothed:Online:NeurIPS}, retaining the optimal
$O(\sqrt{T(1+P_T)})$ dynamic regret with switching cost.
Alongside these dynamic-regret results,
\citet{pmlr-v235-bhuyan24a} seek simultaneous guarantees in
adversarial and stochastic environments.
For quadratic hitting costs and squared $\ell_2$-norm switching
costs, they combine adversarial competitiveness with near-optimal
stochastic performance when the hitting-cost minimizers form
a martingale whose increments have a common covariance matrix. Our focus is on simultaneously controlling the competitive ratio
and dynamic regret against arbitrary comparator sequences in
the lookahead setting.
We achieve this by aggregating experts designed for these two
criteria and bounding the additional cost relative to each expert.

\begin{sidewaystable}[p]
\centering
\caption{Results on Dynamic Regret with or without Switching Cost.
All algorithms assume that $f_t(\cdot)$ is convex and $\X$ is
convex and bounded, so we omit these conditions.}
\label{tab:results:dynamic:regret}
\renewcommand{\arraystretch}{1.3}\begin{tabular}{
@{}c@{\hspace{.5ex}}|
@{\hspace{.5ex}}c@{\hspace{.5ex}}|
@{\hspace{.5ex}}c@{\hspace{.5ex}}|
@{\hspace{.5ex}}c@{}
}
\toprule
Algorithm & \tabincell{c}{Setting and\\Assumptions} & Metric & Upper Bounds \\ \hline
OBD& Lookahead & \multirow{2}{*}{$\displaystyle
\sum_{t=1}^T \big( f_t(\x_t)+\|\x_t-\x_{t-1}\|
-f_t(\u_t)-\|\u_t-\u_{t-1}\| \big)$}
&\multirow{2}{*}{$O\left(\sqrt{TL}\right)$}
\\
\citep{SOCO:OBD} & $P_T\leq L$ & & \\ \hline
\multirow{3}{*}{\tabincell{c}{R-OBD\\\citep[Theorem~6]{NIPS2019_8463}}}
& Lookahead &
\multirow{5}{*}{$\displaystyle \sum_{t=1}^T
\Big(f_t(\x_t)+\frac12\|\x_t-\x_{t-1}\|^2
-f_t(\u_t)-\frac12\|\u_t-\u_{t-1}\|^2
\Big)$}
&\multirow{5}{*}{$O(1+P_T)$} \\
& Vanishing gradient & & \\
& condition & & \\ \cline{1-2}

The Greedy Algorithm  & \multirow{2}{*}{Lookahead} & & \\
(Lemma~\ref{lem:qbow:dynamic:expert}) & & & \\ \hline
OGD & \multirow{6}{*}{
\tabincell{c}{ Standard OCO\\
Bounded gradients
}
}
&
\multirow{4}{*}{$\displaystyle
\sum_{t=1}^T
\big(f_t(\x_t)-f_t(\u_t)\big)$}
&
\multirow{2}{*}{$O\left(\sqrt{T}(1+P_T)\right)$}
\\
\citep{zinkevich-2003-online} & & & \\ \cline{1-1} \cline{4-4}

Ader &  &  &
\multirow{6}{*}{$O\left(\sqrt{T(1+P_T)}\right)$}
\\
\citep{Adaptive:Dynamic:Regret:NIPS} & & & \\ \cline{1-1} \cline{3-3}
SAder & &
\multirow{4}{*}{$\displaystyle
\sum_{t=1}^T
\big(f_t(\x_t)+\|\x_t-\x_{t-1}\|-f_t(\u_t)
\big)$} &
\\
(Theorem~\ref{thm:ader}) & & &
\\ \cline{1-2}
Lookahead SAder & \multirow{2}{*}{Lookahead} &  &
\\
(Theorem~\ref{thm:lookahead:ader}) & & &
\\ \bottomrule
\end{tabular}
\end{sidewaystable}
\subsection{Dynamic Regret} \label{ssec:dynamic_regret}
Recently, dynamic regret has attracted considerable interest in the community of online learning \citep{IJCAI:2020:Zhang}. The motivation of dynamic regret is to deal with changing environments, in which the optimal decision may change over time. It is defined as the difference between the cumulative loss of the learner and that of a sequence of comparators $\u_1, \ldots, \u_T \in \X$:
\begin{equation} \label{eqn:dynamic:1}
\DReg(\u_1,\ldots,\u_T) = \sum_{t=1}^T f_t(\x_t)  - \sum_{t=1}^T f_t(\u_t).
\end{equation}
In the general form of dynamic regret,  $\u_1, \ldots, \u_T$ could be an \emph{arbitrary} sequence \citep{zinkevich-2003-online,Dynamic:ICML:13,Adaptive:Dynamic:Regret:NIPS,Adaptive:Dynamic:AISTATS,pmlr-v119-cutkosky20a,Problem:Dynamic:Regret,pmlr-v134-baby21a,AAAI:2024:Wang,Switching:to:Dynamic}.  In the restricted form, the comparators are chosen as per-round minimizers,  i.e., $\u_t \in \argmin_{\x \in \X} f_t(\x)$ \citep{Dynamic:AISTATS:15,Non-Stationary,Dynamic:2016,Dynamic:Strongly,Dynamic:Regret:Squared,Dynamic:Regret:Adaptive,AAAI:2021:Wan:B,Worst:Dynamic:SSS,NeurIPS:2024:Xu}. While it is well-known that sublinear dynamic regret is unattainable in the worst case, one can bound the dynamic regret in terms of some regularities of the comparator sequence. An instance is given by \citet{zinkevich-2003-online}, who introduces the notion of path-length defined in (\ref{eqn:path}) to measure the temporal variability of the comparator sequence, and derives an $O(\sqrt{T}(1+P_T))$ bound for the dynamic regret of OGD. Later, \citet{Adaptive:Dynamic:Regret:NIPS} develop adaptive learning for dynamic environment (Ader), which achieves the optimal $O(\sqrt{T(1+P_T)})$ dynamic regret. In this paper, we show that a small change of Ader attains the same bound for dynamic regret with switching cost.  We summarize our results and representative previous guarantees in Table~\ref{tab:results:dynamic:regret}.

\subsection{Switching Constraints}

Another line of work studies online convex optimization under
explicit switching constraints.
\citet{NEURIPS2020_236f119f} characterize the minimax regret
when the learner is allowed only a limited number of switches.
\citet{Sherman2023LazyOCO} study a related formulation in which
the switching budget is imposed in expectation.
Subsequent work improves regret guarantees under this expected
switch-count budget
\citep{pmlr-v195-agarwal23d,pmlr-v235-agarwal24d},
with \citet{pmlr-v235-agarwal24d} obtaining better dimension
dependence without requiring smooth losses.
Beyond switch-count constraints,
\citet{NeurIPS2021:OCO:Constraint} impose a budget on the
cumulative movement of the learner and establish minimax-optimal
regret bounds for convex losses.
This formulation accounts for the distance between consecutive
decisions, allowing finer control over movement than counting
each change equally.
These studies minimize hitting-cost regret against the best
fixed decision while satisfying a prescribed switching budget.
In our setting, switching costs are included in the objective
together with hitting costs, so the learner directly trades off
the two types of costs.

\section{Competitive Ratio} \label{sec:ratio}
In this section, we study competitive ratio as the performance metric. Throughout the paper, we assume nonnegative hitting costs whenever competitive ratio is considered, as in prior work  \citep{SOCO:OBD,pmlr-v125-argue20a}. For our competitive-ratio guarantees, the minimization in \eqref{eqn:competitive:ratio} is restricted to comparator
sequences with $\u_0=\x_0$.

\subsection{Polyhedral Functions}
We first introduce the definition of polyhedral functions.
\begin{definition} A function $f(\cdot): \X \to \R$ with minimizer $\bv$ is
$\alpha$-polyhedral if
\begin{equation} \label{eqn:polyhedral}
f(\x)- f(\bv) \geq \alpha \|\x-\bv\|,  \  \forall \x  \in \X.
\end{equation}
\end{definition}
We note that polyhedral functions have been used for stochastic network optimization \citep{stochastic:network:opt} and geographical load balancing \citep{Geographical:Load:Balance}.

Following \citet{SOCO:OBD}, we set the switching cost as $m(\x_t,\x_{t-1})=\|\x_t-\x_{t-1}\|$. Intuitively, we may expect that the switching cost should be taken into consideration when making decisions. However, our analysis shows that minimizing the hitting cost alone yields the tightest competitive ratio so far. Specifically, we consider the following naive approach that ignores the switching cost and selects
\begin{equation} \label{eqn:naive}
\x_t = \argmin_{\x\in \X} f_t(\x).
\end{equation}

The theoretical guarantee of (\ref{eqn:naive}) is stated below.
\begin{thm} \label{thm:ratio:poly} Suppose each $f_t(\cdot): \X \to \R_+$  with minimizer $\bv_t$ is $\alpha$-polyhedral. We have
\begin{equation}\label{eqn:ratio:poly}
\sum_{t=1}^T \big( f_t(\x_t) + \|\x_t-\x_{t-1}\| \big)  \leq  \max\left( 1, \frac{2}{\alpha} \right) \sum_{t=1}^T  \big(f_t (\u_t) + \|\u_t - \u_{t-1}\|\big)
\end{equation}
for every comparator sequence $\u_0, \u_1, \ldots, \u_T \in \X$ with $\x_0=\u_0$. Moreover, the above bound is tight for the naive algorithm, even on the one-dimensional domain $\X=[-1,1]$.
\end{thm}
\textbf{Remark:} Our $\max(1,\frac{2}{\alpha})$ competitive ratio is much better than the $3+\frac{8}{\alpha}$ ratio of OBD
\citep{SOCO:OBD}, and also improves upon the
$1+\frac{2}{\alpha}$ bound established by
\citet{10.1145/3379484} for (\ref{eqn:naive}).
Furthermore, the proof of Theorem~\ref{thm:ratio:poly}
is much simpler than that of OBD and refines the analysis
of \citet[Lemma~1]{10.1145/3379484}.

We next examine the effect of explicitly accounting for
the switching cost through a fixed weight in the greedy
algorithm~\eqref{eqn:greedy}.
\begin{prop}\label{prop:greedy:poly}
Fix $\alpha>0$.
For any fixed $\gamma\geq0$, the worst-case competitive
ratio of the greedy algorithm~\eqref{eqn:greedy},
with the $\ell_2$-norm switching cost, over nonnegative
$\alpha$-polyhedral hitting costs is at least
$\max\left(1,\frac{2}{\alpha}\right)$, regardless of
how ties are broken.
Moreover, when $\gamma<\alpha$, the algorithm coincides
with the naive algorithm, whereas when $\gamma>\alpha$,
its competitive ratio is unbounded.
\end{prop}

For $\alpha\geq2$, Theorem~\ref{thm:ratio:poly} gives a competitive ratio of $1$, which is optimal. For $1\leq\alpha<2$, we establish a matching lower bound for deterministic online algorithms, yielding the following optimality result for all $\alpha\geq1$.
\begin{thm}\label{thm:poly:lower}
For $\alpha\geq1$, the naive algorithm achieves the optimal
competitive ratio $\max\left(1,\frac{2}{\alpha}\right)$
among deterministic online algorithms for nonnegative
$\alpha$-polyhedral hitting costs with the $\ell_2$-norm
switching cost.
\end{thm}
\textbf{Remark:}
For $0<\alpha<1$, the optimal deterministic competitive
ratio remains open.
Proposition~\ref{prop:greedy:poly} shows that any improvement
over the naive guarantee would require going beyond
the greedy algorithm with a fixed switching-cost weight.
\subsection{Quadratic Growth Functions}
In this section, we consider the quadratic growth condition.
\begin{definition} A function $f(\cdot): \X \to \R$ with minimizer $\bv$ is
$\lambda$-quadratic growth if
\begin{equation} \label{eqn:quadratic:growth}
f(\x)- f(\bv) \geq \frac{\lambda}{2} \|\x-\bv\|^2,  \  \forall \x  \in \X.
\end{equation}
\end{definition}
The quadratic growth condition has been exploited by the optimization community \citep{moor.2017.0889,Necoara2019} to establish linear convergence, and this condition is weaker than  strong convexity \citep{COLT:Hazan:2011}.

Following \citet{NIPS2019_8463}, we set the switching cost as $m(\x_t,\x_{t-1})=\|\x_t-\x_{t-1}\|^2/2$. We also consider the naive approach in (\ref{eqn:naive}) and have the following theoretical guarantee.
\begin{thm} \label{thm:naive:quadratic} Suppose each $f_t(\cdot): \X \to \R_+$  with minimizer $\bv_t$ is $\lambda$-quadratic growth. We have
\[
\begin{split}\sum_{t=1}^T \left( f_t(\x_t) + \frac{1}{2}\|\x_t-\x_{t-1}\|^2 \right)  \leq \left( 1 + \frac{4}{\lambda}\right) \sum_{t=1}^T  \left(f_t (\u_t) + \frac{1}{2} \|\u_t - \u_{t-1}\|^2\right),
 \end{split}
\]
for every comparator sequence $\u_0, \u_1, \ldots, \u_T \in \X$ with $\x_0=\u_0$.
\end{thm}
\textbf{Remark:}  The above theorem implies that the naive approach achieves a competitive ratio of $1+\frac{4}{\lambda}$,
which matches the order of the lower bound for this algorithm as $\lambda$ tends to zero \citep[Theorem~5]{NIPS2019_8463}.  Furthermore, it is also much better than the
$\max(1+\frac{6}{\lambda},4)$ ratio established by
\citet{10.1145/3379484} for (\ref{eqn:naive}).
Whether explicitly accounting for the switching cost
can improve this guarantee under the quadratic growth
condition alone remains open.

\subsection{Convex and Quadratic Growth Functions}
When $f_t(\cdot)$ is both quasiconvex and $\lambda$-quadratic growth, \citet{NIPS2019_8463} have established an $O(1/\sqrt{\lambda})$ competitive ratio for Greedy OBD. Inspired by this result, we introduce convexity to further improve the competitive ratio. In this case, the switching cost plays a role in deriving tighter competitive ratios.  Specifically, we choose the greedy algorithm with $\gamma >0$ to select $\x_t$, i.e.,
\begin{equation} \label{eqn:2}
\x_t = \argmin_{\x\in \X}\left(  f_t(\x)+ \frac{\gamma}{2} \|\x-\x_{t-1}\|^2 \right).
\end{equation}
The theoretical guarantee of (\ref{eqn:2}) is stated below.

\begin{thm} \label{thm:greedy:convex:quadratic} Suppose the domain $\X$ is convex, and each $f_t(\cdot): \X \to \R_+$  with minimizer $\bv_t$ is convex and $\lambda$-quadratic growth. By setting
\[
\gamma=\frac{\lambda}{\lambda + \sqrt{\lambda}},
\]
 we have
\begin{equation}\label{eqn:greedy:convex:quadratic}
\sum_{t=1}^T \left( f_t(\x_t) + \frac{1}{2}\|\x_t-\x_{t-1}\|^2 \right) \leq  \left( 1 + \frac{2}{\sqrt{\lambda}}\right) \sum_{t=1}^T  \left(f_t (\u_t) + \frac{1}{2} \|\u_t - \u_{t-1}\|^2\right),
\end{equation}
for every comparator sequence $\u_0, \u_1, \ldots, \u_T \in \X$ with $\x_0=\u_0$.
\end{thm}
\textbf{Remark:}  Under the additional convexity assumption, accounting for switching costs improves the competitive-ratio bound from
$1+\frac{4}{\lambda}$ to $1+\frac{2}{\sqrt{\lambda}}$ when $\lambda$ is small. According to the $\frac{1}{2}+ \frac{1}{2}\sqrt{1+\frac{4}{\lambda}}$ lower bound of the competitive ratio of strongly convex functions \citep[Theorem 1]{NIPS2019_8463}, the ratio in Theorem~\ref{thm:greedy:convex:quadratic} is optimal up to constant factors.  Compared with Greedy OBD \citep[Theorem 3]{NIPS2019_8463}, our assumption is slightly stronger, since we require convexity instead of quasiconvexity. However, our algorithm and analysis are much simpler, and the constants in our bound are much smaller.

\section{Dynamic Regret with Switching Cost} \label{sec:regret}
When considering dynamic regret with switching cost, we adopt the common assumptions of online convex optimization (OCO) \citep{Online:suvery}.
\begin{ass}\label{ass:0} All functions $f_t$ are convex over their domain $\X$.
\end{ass}
\begin{ass}\label{ass:2} The domain $\X$ is convex and has bounded diameter $D$, i.e.,
\begin{equation}\label{eqn:domain}
\max_{\x, \x' \in \X} \|\x -\x'\| \leq D.
\end{equation}
\end{ass}
\begin{ass}\label{ass:1} The gradients of all functions are bounded by $G$, i.e.,
\begin{equation}\label{eqn:grad}
\max_{\x \in \X}\|\nabla f_t(\x)\| \leq G, \ \forall t \in [T].
\end{equation}
\end{ass}
Assumption~\ref{ass:1} implies that the hitting cost is Lipschitz continuous, so it is natural to set the switching cost as $m(\x_t,\x_{t-1})=\|\x_t-\x_{t-1}\|$.

\subsection{The Standard Setting}
We first follow the standard setting of OCO in which the learner cannot observe the hitting cost when making predictions, and develop an algorithm based on Ader \citep{Adaptive:Dynamic:Regret:NIPS},  which is designed to minimize the dynamic regret w.r.t.~any comparator sequence. Ader runs multiple OGD instances with different step sizes as expert-algorithms, and uses Hedge as the meta-algorithm to aggregate predictions from experts. In Ader, the meta-algorithm uses the hitting cost $f_t(\cdot)$ or its linearized variant to measure the performance of experts. Specifically, we demonstrate that a small modification to Ader, which modifies the loss of the meta-algorithm to take into account the switching cost of experts, is sufficient to minimize the dynamic regret with switching cost. We call the proposed method Smoothed Ader (SAder) and describe it below.

\begin{algorithm}[t]
\caption{SAder: Meta-algorithm}
\begin{algorithmic}[1]
\REQUIRE An initial point $\x_0\in\X$, a step size $\beta$, and a set $\H$ of expert step sizes
\STATE For each $\eta\in\H$, activate expert $E^\eta$ by invoking Algorithm~\ref{alg:2} with $\eta$ and $\x_0$
\STATE Sort step sizes in ascending order $\eta_1 \leq \eta_2 \leq \cdots \leq \eta_N$, and set $w_1^{\eta_i}=\frac{C}{i(i+1)}$
\FOR{$t=1,\ldots,T$}
\STATE Receive $\x_t^\eta$ from each expert $E^\eta$
\STATE Output the weighted average $\x_t = \sum_{\eta \in \H} w_t^\eta \x_t^\eta$
\STATE Observe the loss function $f_t(\cdot)$
\STATE Update the weight of each expert by (\ref{eqn:meta:update})
\STATE Send gradient $\nabla f_t(\x_t)$ to each expert $E^\eta$
\ENDFOR
\end{algorithmic}
\label{alg:1}
\end{algorithm}

\paragraph{Meta-algorithm} The meta-algorithm is similar to that of Ader \citep[Algorithm 3]{Adaptive:Dynamic:Regret:NIPS}, and summarized in  Algorithm~\ref{alg:1}. The meta-algorithm takes as input an arbitrary initial point $\x_0\in\X$, a step size $\beta$, and a set $\H$ of expert step sizes.  In Step 1, it
activates each expert $E^\eta$ by invoking Algorithm~\ref{alg:2} with $\eta$ and $\x_0$.  In Step 2, we set the initial weight of each expert. Let $\eta_i$ be the $i$-th smallest step size in $\H$. The weight of $E^{\eta_i}$ is chosen as
\begin{equation}\label{eqn:initial:weight}
w_1^{\eta_i}=\frac{C}{i(i+1)}, \textrm{ and } C=1+\frac{1}{|\H|}.
\end{equation}
In each round, the meta-algorithm receives a set of predictions $\{\x_t^\eta|\eta \in \H\}$ from all experts (Step 4), and outputs the weighted average (Step 5):
\[
\x_t = \sum_{\eta \in \H} w_t^\eta \x_t^\eta
\]
where $w_t^\eta$ is the weight assigned to expert $E^\eta$. After observing the loss function, the weights of experts are updated according to the  exponential weighting scheme (Step 7) \citep{bianchi-2006-prediction}:
\begin{equation} \label{eqn:meta:update}
w_{t+1}^\eta = \frac{w_{t}^\eta e^{-\beta \ell_t(\x_t^\eta)} }{\sum_{\eta \in \H} w_{t}^\eta e^{-\beta \ell_t(\x_t^\eta)}}
\end{equation}
where
\begin{equation} \label{eqn:meta:loss:function}
\ell_t(\x_t^\eta) = \langle \nabla f_t(\x_t), \x_t^\eta -\x_t \rangle + \|\x_t^\eta - \x_{t-1}^\eta\|.
\end{equation}
At $t=0$, all experts are initialized at the same point as the meta-algorithm; that is, $\x_0^\eta=\x_0$ for every $\eta\in\H$. As can be seen from (\ref{eqn:meta:loss:function}), we incorporate the switching cost $\|\x_t^\eta - \x_{t-1}^\eta\|$ of expert $E^\eta$ to measure its performance. This is the \emph{only} modification made to Ader. In the last step, we send the gradient $\nabla f_t(\x_t)$ to each expert $E^\eta$ so that they can update their own predictions.

\begin{algorithm}[t]
\caption{SAder: Expert-algorithm}
\begin{algorithmic}[1]
\REQUIRE The step size $\eta$ and an initial point $\x_0\in\X$
\STATE Set $\x_0^\eta=\x_1^\eta=\x_0$
\FOR{$t=1,\ldots,T$}
\STATE Submit $\x_t^\eta$ to the meta-algorithm
\STATE Receive gradient $\nabla f_t(\x_t)$ from the meta-algorithm
\STATE
\[
\x_{t+1}^\eta= \Pi_{\X}\big[\x_t^\eta - \eta \nabla f_t(\x_t)\big]
\]
\ENDFOR
\end{algorithmic}
\label{alg:2}
\end{algorithm}
\paragraph{Expert-algorithm} The expert-algorithm is the same as that of Ader \citep[Algorithm 4]{Adaptive:Dynamic:Regret:NIPS}, which is OGD over the linearized loss
\begin{equation} \label{eqn:surrogate}
s_t(\x)=\langle \nabla f_t(\x_t), \x -\x_t \rangle .
\end{equation}
For completeness, we present its procedure in Algorithm~\ref{alg:2}. The inputs to each expert are its step size $\eta$ and the common initial point $\x_0$. In Step 3, each expert submits its prediction $\x_t^\eta$ to the meta-algorithm, and receives the gradient $\nabla f_t(\x_t)$ in Step 4. Then, in Step 5, it performs gradient descent
\[
\x_{t+1}^\eta= \Pi_{\X}\big[\x_t^\eta - \eta \nabla f_t(\x_t)\big]
\]
to get the prediction for the next round. Here, $\Pi_{\X}[\cdot]$ denotes the projection onto the nearest point in $\X$.

We have the following theoretical guarantee.
\begin{thm} \label{thm:ader} Set
\begin{equation} \label{eqn:def:H}
\H=\left\{ \left. \eta_i=2^{i-1} \sqrt{\frac{D^2}{T (G^2+2G)}} \right  | i=1,\ldots, N\right\}
\end{equation}
where
\[
N= \left\lceil \frac{1}{2} \log_2 (1+2T) \right\rceil +1, \textrm{ and set } \beta= \frac{2}{(2G+1) D} \sqrt{\frac{2 }{5 T}}
\]
in Algorithm~\ref{alg:1}. Under Assumptions~\ref{ass:0}, \ref{ass:2} and \ref{ass:1},  for any initial point $\x_0\in\X$ and every comparator sequence $\u_0, \u_1,\ldots,\u_T \in \X$, SAder satisfies
\begin{equation}\label{eqn:sader:metric}
\begin{split} 
& \sum_{t=1}^T \big( f_t(\x_t) + \|\x_t -\x_{t-1}\|\big)-  \sum_{t=1}^T  f_t(\u_t)  \\
\leq & \frac{3}{2}\sqrt{T (G^2+2G)\left(D^2+ 2 D  \sum_{t=1}^T  \|\u_t - \u_{t-1}\| \right)}+ (2G+1) D\sqrt{\frac{5  T }{8}} \left[1+ 2\ln (k+1)\right]  \\
=&O\left(\sqrt{T(1+P_T)} + \sqrt{T}(1+\log \log P_T)\right)=O\left(\sqrt{T(1+P_T)} \right) 
\end{split}
\end{equation}
where 
\begin{equation} \label{eqn:def:k}
k = \left\lfloor \frac{1}{2} \log_2 \left(1+ \frac{2P_T}{D} \right) \right \rfloor+1.
\end{equation}
\end{thm}
\textbf{Remark:}  Theorem~\ref{thm:ader} yields an $O(\sqrt{T(1+P_T)})$ bound for dynamic regret with switching cost, even without subtracting the comparator's switching cost. This stronger guarantee in  \eqref{eqn:sader:metric} matches Ader's rate and is optimal up to constant factors by the lower bound for ordinary dynamic regret \citep[Theorem~2]{Adaptive:Dynamic:Regret:NIPS}. Compared with the regret bound of OBD in (\ref{eqn:obd:regret}) \citep{SOCO:OBD}, the advantage of SAder is that its regret depends directly on the path-length $P_T$, and thus becomes tighter for comparator sequences with smaller path-lengths.

\begin{algorithm}[t]
\caption{Lookahead SAder: Meta-algorithm}
\begin{algorithmic}[1]
\REQUIRE An initial point $\x_0\in\X$, a step size $\beta$, and a set $\H$ of expert step sizes
\STATE For each $\eta\in\H$, activate expert $E^\eta$ by invoking Algorithm~\ref{alg:4} with $\eta$ and $\x_0$
\STATE Sort step sizes in ascending order $\eta_1 \leq \eta_2 \leq \cdots \leq \eta_N$, and set $w_0^{\eta_i}=\frac{C}{i(i+1)}$
\FOR{$t=1,\ldots,T$}
\STATE Observe the loss function $f_t(\cdot)$ and send it to each expert $E^\eta$
\STATE Receive $\x_t^\eta$ from each expert $E^\eta$
\STATE Update the weight of each expert by (\ref{eqn:lookahead:hedge:pro})
\STATE Output the weighted average $\x_t = \sum_{\eta \in \H} w_t^\eta \x_t^\eta$
\ENDFOR
\end{algorithmic}
\label{alg:3}
\end{algorithm}

\subsection{The Lookahead Setting}
We next turn to the lookahead setting and investigate whether access to the current hitting cost can yield a stronger guarantee. We show that a lookahead version of SAder achieves a regret bound of the same order while removing Assumption~\ref{ass:1}; in particular, neither the gradients nor the hitting costs need to be bounded.

\paragraph{Meta-algorithm} We design a lookahead version of Hedge, and summarize it in Algorithm~\ref{alg:3}. Compared with Algorithm~\ref{alg:1}, we make the following modifications.
\begin{compactitem}
\item In the $t$-th round, the meta-algorithm first sends $f_t(\cdot)$ to all experts so that they can also benefit from the prior knowledge of $f_t(\cdot)$ (Step 4).
\item After receiving the predictions from the experts (Step 5), the meta-algorithm makes use of $f_t(\cdot)$ to determine the weights of experts (Step 6):
\begin{equation} \label{eqn:lookahead:hedge:pro}
w_{t}^\eta = \frac{w_{t-1}^\eta e^{-\beta c_t(\x_{t}^\eta)} }{\sum_{\eta \in \H} w_{t-1}^\eta e^{-\beta c_t(\x_t^\eta)}},
\end{equation}
where 
\begin{equation}\label{eqn:lookahead:loss}
c_t(\x_t^\eta)=f_t(\x_t^\eta)+\|\x_t^\eta-\x_{t-1}^\eta\|.
\end{equation}
The conference version \citep{Smoothed:Online:NeurIPS} used the linearized loss in (\ref{eqn:meta:loss:function}) for this update, whereas we use the current one-round cost in (\ref{eqn:lookahead:loss}) to avoid the resulting circular dependence between $\x_t$ and the expert weights.
\end{compactitem}

\paragraph{Expert-algorithm} To exploit the hitting cost of the current round, we use the quadratic-regularized greedy update in (\ref{eqn:2})
with $\gamma=1/\eta$.  
The inputs to each expert are its step size $\eta$ and the common initial point $\x_0$. After receiving $f_t(\cdot)$ (Step 3), the expert solves the following optimization problem to obtain $\x_t^\eta$ (Step 4):
\begin{equation} \label{eqn:lookhead:expert}
\min_{\x\in \X} \quad f_t(\x)+ \frac{1}{2 \eta} \|\x -\x_{t-1}^\eta\|^2 .
\end{equation}

\begin{algorithm}[t]
\caption{Lookahead SAder: Expert-algorithm}
\begin{algorithmic}[1]
\REQUIRE The step size $\eta$ and an initial point $\x_0\in\X$
\STATE Set $\x_0^\eta=\x_0$
\FOR{$t=1,\ldots,T$}
\STATE Receive the loss $f_t(\cdot)$ from the meta-algorithm
\STATE Solve the optimization problem in (\ref{eqn:lookhead:expert}) to obtain $\x_t^\eta$
\STATE Submit $\x_t^\eta$ to the meta-algorithm
\ENDFOR
\end{algorithmic}
\label{alg:4}
\end{algorithm}
We have the following theoretical guarantee of the lookahead SAder.
\begin{thm} \label{thm:lookahead:ader} Set
\begin{equation} \label{eqn:def:H:New}
\H=\left\{ \left. \eta_i=2^{i-1} \sqrt{\frac{D^2}{T}} \right  | i=1,\ldots, N\right\}
\end{equation}
where
\[
N= \left\lceil \frac{1}{2} \log_2 (1+2T) \right\rceil +1, \textrm{ and set } \beta=  \frac{1}{D}  \sqrt{\frac{2}{T}}
\]
in Algorithm~\ref{alg:3}. Under Assumptions~\ref{ass:0} and \ref{ass:2},  for any initial point $\x_0\in\X$ and every comparator sequence $\u_0, \u_1,\ldots,\u_T \in \X$, the lookahead SAder satisfies
\begin{equation} \label{eqn:look:sader}
\begin{split}
& \sum_{t=1}^T \big( f_t(\x_t) + \|\x_t -\x_{t-1}\|\big)-  \sum_{t=1}^T  f_t(\u_t)\\
\leq & \frac{3}{2}\sqrt{T (D^2+ 2 D \sum_{t=1}^T  \|\u_t - \u_{t-1}\|)}+ D \sqrt{ \frac{T}{2}}  \left[1+ 2\ln (k+1)\right]\\
=&O\left(\sqrt{T(1+P_T)} + \sqrt{T}(1+\log \log P_T)\right)=O\left(\sqrt{T(1+P_T)} \right)
\end{split}
\end{equation}
where $k$ is defined in (\ref{eqn:def:k}).
\end{thm}
\textbf{Remark:} Similar to SAder, lookahead SAder also achieves an
$O(\sqrt{T(1+P_T)})$ bound for dynamic regret with switching
cost. In the lookahead setting, we no longer require
Assumption~\ref{ass:1}, and the constants in
Theorem~\ref{thm:lookahead:ader} are independent of $G$.
The analysis does not require differentiability of the
hitting costs either: the expert bound in
Lemma~\ref{lem:lookhead:expert} is established using the
strong convexity of the experts' regularized objectives.

To show the optimality of Theorem~\ref{thm:lookahead:ader}, we provide the lower bound of dynamic regret with switching cost under the lookahead setting.
\begin{thm} \label{thm:lower} For any deterministic online algorithm with lookahead ability and any $\tau \in [0, TD]$, there exists a sequence of functions $f_1,\ldots,f_T$ and a sequence of comparators $\u_0,\ldots,\u_T$ satisfying Assumptions~\ref{ass:0} and \ref{ass:2} such that (i) the path-length of  $\u_0,\ldots,\u_T$ is at most $\tau$ and (ii) the dynamic regret with switching cost relative to this comparator sequence is at least $\Omega(\sqrt{T (D^2 + D \tau)})$.
\end{thm}
\textbf{Remark:} The above theorem indicates an $\Omega(\sqrt{T(1+P_T)})$ lower bound, which implies that the lookahead SAder is optimal up to constant factors. Thus, even in the lookahead setting, the $O(\sqrt{T(1+P_T)})$ upper bound cannot be improved. 

\section{Best of Both Worlds} \label{sec:bow}
In the preceding sections, we have studied the competitive ratio and dynamic regret with switching cost separately.
We now investigate whether a single algorithm can simultaneously achieve strong guarantees for both performance measures.
We first consider the $\ell_2$-norm switching cost and then extend our approach to its squared counterpart.

\subsection{Technical Challenges}
For the $\ell_2$-norm switching cost $m(\x_t,\x_{t-1})=\|\x_t-\x_{t-1}\|$, \citet{pmlr-v98-daniely19a} developed a general framework for combining
competitive-ratio and regret guarantees in the presence of switching costs. Following this principle, we consider two experts, designed respectively for competitive ratio and dynamic regret with switching cost, and aggregate their predictions using the Discounted-Normal-Predictor
(DNP) \citep{DNP:2010:Arxiv,NIPS2011_11b921ef}. Even in the presence of switching costs, DNP guarantees constant regret w.r.t.~the first expert and $\Ot(\sqrt{T})$ regret w.r.t.~the second expert. Consequently, the resulting meta-algorithm preserves the competitive-ratio guarantee of the former up to a constant additive term while adding only an $\Ot(\sqrt{T})$ term to the dynamic-regret bound of the latter.

Directly applying this methodology requires a known bound on the one-round cost difference between the two experts to normalize the
feedback supplied to DNP.  On a bounded domain, the bounded-gradient Assumption~\ref{ass:1} would provide such a bound.  However, the naive algorithm and lookahead SAder achieve their respective guarantees without assuming bounded gradients (Theorems~\ref{thm:ratio:poly} and~\ref{thm:lookahead:ader}). We therefore seek to aggregate these two experts without introducing this additional assumption.  The lookahead protocol provides a way to overcome this difficulty: after observing the current hitting cost and obtaining the predictions of the two experts, the meta-algorithm knows their current cost difference before making its own decision. This additional information motivates a lookahead-aware variant of DNP that operates without a prescribed bound on the experts' cost differences.

\subsection{$\ell_2$-Norm Switching Cost} \label{sec:bow:l2}
We consider two arbitrary experts, denoted by $E^0$ and $E^1$,
and initialize both at a common point $\x_0\in\X$.
After observing $f_t(\cdot)$ at round $t$, the two experts
produce candidate decisions $\x_t^0$ and $\x_t^1$, respectively.
Define the one-round cost of expert $E^i$ by
 \begin{equation}\label{eqn:bow:expert:cost}
	C_t^i = f_t(\x_t^i) + \|\x_t^i - \x_{t-1}^i\|,  \   i\in\{0,1\} .
\end{equation}
Let $\Delta_t=C_t^0-C_t^1$ denote their cost difference. The meta-algorithm outputs the convex combination
\begin{equation}\label{eqn:bow:prediction}
 	\x_t = (1-p_t)\x_t^0 + p_t\x_t^1, 
\end{equation}
where $p_t\in[0,1]$ is the mixing weight, and we set $p_0=0$.

Changing the mixing weight also incurs an additional switching cost. The following lemma characterizes the cumulative cost of the
meta-algorithm by relating it to the costs of the two experts.
\begin{lem}\label{lem:bow:interpolation}
Under Assumptions~\ref{ass:0} and~\ref{ass:2}, the cumulative cost of the meta-algorithm satisfies
\begin{equation} \label{eq:bow:interpolation:E0} 
\sum_{t=1}^{T} 
\big(f_t(\x_t)+\|\x_t-\x_{t-1}\| \big) \leq \sum_{t=1}^{T}C_t^0 + \sum_{t=1}^{T} \big(-p_t\Delta_t + D|p_t-p_{t-1}| \big),
\end{equation}
and
\begin{equation} \label{eq:bow:interpolation:E1}
\sum_{t=1}^{T}
\big(f_t(\x_t)+\|\x_t-\x_{t-1}\| \big)  \leq \sum_{t=1}^{T}C_t^1+ \sum_{t=1}^{T} \big((1-p_t)\Delta_t +D|p_t-p_{t-1}| \big).
\end{equation}
\end{lem}

Lemma~\ref{lem:bow:interpolation} reduces the analysis to controlling
the two residual terms involving $\Delta_t$ and $|p_t-p_{t-1}|$.
To handle these terms without a prescribed bound on the cost differences,
we introduce UDNP, an unbounded-input variant of DNP.
UDNP follows the mixing weight supplied by DNP when $|\Delta_t|<2D$
and selects the expert with the smaller current cost otherwise.
The procedure is summarized in Algorithm~\ref{alg:udnp}.

Before observing $f_t(\cdot)$, the internal DNP instance outputs a reference weight $q_t\in[0,1]$ based on past feedback.
After obtaining the current expert predictions, UDNP evaluates the cost difference $\Delta_t$ and sets
\begin{equation}\label{eqn:bow:weight}
	p_t =
	\begin{cases}
		1, & \Delta_t\geq 2D,\\
		0, & \Delta_t\leq -2D,\\
		q_t, & |\Delta_t|<2D.
	\end{cases}
\end{equation}
The threshold $2D$ ensures that the gain from overriding $q_t$ compensates for the resulting increase in the cumulative switching penalty.

After choosing $p_t$, UDNP constructs bounded feedback for the internal DNP instance by clipping and normalizing the current cost difference:
\begin{equation}\label{eqn:bow:clipping}
	\dtb_t = \Pi_{[-2D,2D]}[\Delta_t]  \textrm{ and } b_t = \frac{\dtb_t}{2D}
, \end{equation}
where $\Pi_{[-2D,2D]}[\cdot]$ denotes projection onto the interval $[-2D,2D]$.
The normalized feedback $b_t\in[-1,1]$ is then sent to DNP,
which uses it to update its internal state for the next round.
For completeness, Algorithm~\ref{alg:dnp} gives the internal DNP procedure.
It maintains a discounted state $\theta_t$ and maps it to a reference mixing weight through $q_t=g(\theta_t)$, where
\[
	g(x) = \Pi_{[0,1]}[\tg(x)],
	\ 
	\tg(x) = \sqrt{\frac{T}{8}} Z  \erf\left(\frac{x}{\sqrt{8T}}\right) \exp\left(\frac{x^2}{16T}\right),
	\ 
	\erf(x) = \int_0^x \exp\left(-\frac{s^2}{2}\right) \, ds,
\]
and $Z>0$ is a parameter.
With $\theta_1=0$ and $\rho=1-1/T$, DNP first sends $q_t=g(\theta_t)$ to UDNP,
then receives the normalized feedback $b_t$,
and finally updates its state according to $\theta_{t+1}=\rho\theta_t+b_t$.

\begin{algorithm}[t]
\caption{UDNP: A Meta-Algorithm for the Best of Both Worlds}
\begin{algorithmic}[1]
	\REQUIRE Experts $E^0$ and $E^1$ initialized at a common $\x_0\in\X$, and a DNP instance with parameter $Z$
	\FOR{$t=1,\ldots,T$}
		\STATE Receive $q_t$ from DNP
		\STATE Observe $f_t(\cdot)$ and obtain $\x_t^0$ and $\x_t^1$
		from $E^0$ and $E^1$, respectively
		\STATE Compute $C_t^0$ and $C_t^1$ by
		\eqref{eqn:bow:expert:cost} or \eqref{eq:bow:squared_l2:expert:cost}, and set
		$\Delta_t=C_t^0-C_t^1$
		\STATE Set $p_t$ according to \eqref{eqn:bow:weight} or \eqref{eqn:bow:squared_weight}, and output $\x_t$ according to \eqref{eqn:bow:prediction}
		\STATE Set
		$b_t$ according to \eqref{eqn:bow:clipping} or \eqref{eqn:bow:squared_clipping}
		\STATE Send $b_t$ to DNP
	\ENDFOR
\end{algorithmic}
\label{alg:udnp}
\end{algorithm}

\begin{algorithm}[t]
\caption{Discounted-Normal-Predictor (DNP)}
\begin{algorithmic}[1]
	\REQUIRE Parameter $Z>0$
	\STATE Set $\theta_1=0$ and $\rho=1-1/T$
	\FOR{$t=1,\ldots,T$}
		\STATE Set $q_t=g(\theta_t)$ and send it to UDNP
		\STATE Receive $b_t\in[-1,1]$ from UDNP
		\STATE Set $\theta_{t+1}=\rho\theta_t+b_t$
	\ENDFOR
\end{algorithmic}
\label{alg:dnp}
\end{algorithm}

The following theorem gives the general aggregation guarantee
of UDNP.
\begin{thm}\label{thm:bow}
Suppose Assumptions~\ref{ass:0} and~\ref{ass:2} hold
and $T\geq\max\{8e,16\log T\}$. For any $\x_0\in\X$, run Algorithm~\ref{alg:udnp}
with $Z=1/T$ and any two experts $E^0$ and $E^1$
initialized at $\x_0$, using the weights and feedback
defined in \eqref{eqn:bow:weight} and
\eqref{eqn:bow:clipping}.
Then
\begin{equation}\label{eqn:bow:aggregation}
\sum_{t=1}^T
\left(f_t(\x_t)+\|\x_t-\x_{t-1}\|\right)\leq\min\left\{
\sum_{t=1}^T C_t^0+2D,\;
\sum_{t=1}^T C_t^1+16D\sqrt{T\log T}+6D
\right\}.
\end{equation}
\end{thm}
Theorem~\ref{thm:bow} preserves any competitive-ratio
guarantee of $E^0$ up to an additive constant, while
adding only an $O(\sqrt{T\log T})$ term to the
dynamic-regret bound of $E^1$.
The following corollary instantiates this guarantee
with the naive algorithm and lookahead SAder.
\begin{cor}\label{cor:bow}
Under the conditions of Theorem~\ref{thm:bow},
suppose each $f_t(\cdot):\X\to\R_+$ with minimizer
$\bv_t$ is $\alpha$-polyhedral.
Choose $E^0$ as the naive algorithm in \eqref{eqn:naive}
and $E^1$ as lookahead SAder configured as in
Theorem~\ref{thm:lookahead:ader}. Then
\[
\sum_{t=1}^T
\left(f_t(\x_t)+\|\x_t-\x_{t-1}\|\right)
\leq
\max\left(1,\frac{2}{\alpha}\right)
\sum_{t=1}^T
\left(f_t(\u_t)+\|\u_t-\u_{t-1}\|\right)
+2D,
\]
for every comparator sequence $\u_0,\ldots,\u_T\in\X$
with $\u_0=\x_0$. Moreover, for every $\u_0,\ldots,\u_T\in\X$,
\begin{align*}
&\sum_{t=1}^T
\left(f_t(\x_t)+\|\x_t-\x_{t-1}\|\right)
-\sum_{t=1}^T f_t(\u_t)\displaybreak[1] \\ 
\leq& \frac{3}{2}
\sqrt{
T\left(D^2+2D\sum_{t=1}^T\|\u_t-\u_{t-1}\|\right)
}
+D\sqrt{\frac{T}{2}}\left[1+2\ln(k+1)\right]
+16D\sqrt{T\log T}+6D \\
=&O\left(\sqrt{T(1+P_T)}+\sqrt{T\log T}\right),
\end{align*}
where $k$ is defined in \eqref{eqn:def:k}.
\end{cor}
\textbf{Remark:} Corollary~\ref{cor:bow} preserves the naive algorithm's competitive
factor $\max(1,2/\alpha)$ up to an additive constant $2D$, while
achieving an $O(\sqrt{T(1+P_T)}+\sqrt{T\log T})$ bound for dynamic
regret with switching cost. Both guarantees hold without the
bounded-gradient Assumption~\ref{ass:1}. When $P_T=\Omega(\log T)$,
the regret bound matches that of lookahead SAder up to constant factors.

\subsection{Squared $\ell_2$-Norm Switching Cost} \label{sec:bow:l2square}
We now consider the squared $\ell_2$-norm switching cost
\[
    m(\x_t,\x_{t-1})=\frac{1}{2}\|\x_t-\x_{t-1}\|^2.
\]

We again consider two arbitrary experts, $E^0$ and $E^1$,
initialized at a common point $\x_0\in\X$. After observing $f_t(\cdot)$ at round $t$, the two experts produce candidate decisions $\x_t^0$ and $\x_t^1$, respectively.
Define the one-round cost of expert $E^i$ by
\begin{equation}\label{eq:bow:squared_l2:expert:cost}
    C_t^i=f_t(\x_t^i)  +\frac12\|\x_t^i-\x_{t-1}^i\|^2,     \  i\in\{0,1\}.
\end{equation}
Let $\Delta_t=C_t^0-C_t^1$ denote their cost difference. The meta-algorithm uses the convex combination in \eqref{eqn:bow:prediction}, with $p_t\in[0,1]$ and $p_0=0$. Since the squared switching cost does not satisfy the triangle inequality, the argument underlying Lemma~\ref{lem:bow:interpolation} does not apply directly.
Using the convexity of the squared norm, we obtain
the following counterpart.
\begin{lem}\label{lem:qbow:interpolation}
Under Assumptions~\ref{ass:0} and~\ref{ass:2}, the cumulative cost of the meta-algorithm satisfies
\begin{equation}\label{eq:qbow:interpolation:E0}
\sum_{t=1}^{T}
\left(f_t(\x_t)+\frac12\|\x_t-\x_{t-1}\|^2\right) \leq
\sum_{t=1}^{T}C_t^0 +\sum_{t=1}^{T}
\left(-p_t\Delta_t+\frac{D^2}{2}|p_t-p_{t-1}|\right),
\end{equation}
and
\begin{equation}\label{eq:qbow:interpolation:E1}
\sum_{t=1}^{T} \left(f_t(\x_t)+\frac12\|\x_t-\x_{t-1}\|^2\right) \leq
\sum_{t=1}^{T}C_t^1 +\sum_{t=1}^{T}
\left((1-p_t)\Delta_t+\frac{D^2}{2}|p_t-p_{t-1}|\right).
\end{equation}
\end{lem}

Lemma~\ref{lem:qbow:interpolation} shows that the additional penalty remains linear in the variation of the mixing weights.
The two residual terms have the same form as those in Section~\ref{sec:bow:l2}, with the variation coefficient $D$
replaced by $D^2/2$. We therefore reuse Algorithm~\ref{alg:udnp}, replacing
the threshold and normalization scale $2D$ with $D^2$. Specifically, after receiving the reference weight $q_t$ from DNP and evaluating the current cost difference $\Delta_t$,
UDNP sets
\begin{equation} \label{eqn:bow:squared_weight}
	p_t
	=
	\begin{cases}
		1, & \Delta_t\geq D^2,\\
		0, & \Delta_t\leq-D^2,\\
		q_t, & |\Delta_t|<D^2.
	\end{cases}
\end{equation}
The feedback sent to DNP is obtained by clipping and normalizing the cost difference:
\begin{equation} \label{eqn:bow:squared_clipping}
	\dtb_t=\Pi_{[-D^2,D^2]}[\Delta_t],   \textrm{ and }  b_t=\frac{\dtb_t}{D^2}
. \end{equation}
The internal DNP update remains unchanged.

The following theorem gives the corresponding aggregation
guarantee for the squared $\ell_2$-norm switching cost.
\begin{thm}\label{thm:bow:quadratic}
Suppose Assumptions~\ref{ass:0} and~\ref{ass:2} hold
and $T\geq\max\{8e,16\log T\}$.
For any $\x_0\in\X$, run Algorithm~\ref{alg:udnp}
with $Z=1/T$ and any two experts $E^0$ and $E^1$
initialized at $\x_0$, using the weights and feedback
defined in \eqref{eqn:bow:squared_weight} and
\eqref{eqn:bow:squared_clipping}.
Then
\begin{equation}\label{eqn:qbow:aggregation}
\sum_{t=1}^T
\left(\!f_t(\x_t)+\frac12\|\x_t-\x_{t-1}\|^2\!\right)
\leq
\min\left\{\!
\sum_{t=1}^T C_t^0+D^2,\;
\sum_{t=1}^T C_t^1+8D^2\sqrt{T\log T}+3D^2\!
\right\}.
\end{equation}
\end{thm}
Like Theorem~\ref{thm:bow}, Theorem~\ref{thm:bow:quadratic} applies to arbitrary pairs
of experts. It preserves any competitive-ratio guarantee
of $E^0$ up to an additive constant, while adding only
an $O(\sqrt{T\log T})$ term to the dynamic-regret bound
of $E^1$.

For the competitive-ratio guarantee, we may choose $E^0$
as the naive algorithm in \eqref{eqn:naive} under the
$\lambda$-quadratic growth condition
(Theorem~\ref{thm:naive:quadratic}), the algorithm in
\eqref{eqn:2} under convexity and the $\lambda$-quadratic
growth condition
(Theorem~\ref{thm:greedy:convex:quadratic}),
or R-OBD under $\lambda$-strong convexity
\citep[Theorem~7]{NIPS2019_8463}. For the dynamic-regret guarantee, we may choose $E^1$
as the algorithm in \eqref{eqn:2} with $\gamma=1$,
corresponding to R-OBD with $\lambda_1=1$ and $\lambda_2=0$. 
The proof of  \citet[Theorem~6]{NIPS2019_8463}
uses a vanishing gradient condition for the regularized
objective, which need not hold at a constrained minimizer.
In Lemma~\ref{lem:qbow:dynamic:expert}, we establish
a dynamic-regret bound under Assumptions~\ref{ass:0}
and~\ref{ass:2}, allowing nonsmooth convex hitting costs
and constrained minimizers.

The following corollary instantiates
Theorem~\ref{thm:bow:quadratic} with two greedy experts.
\begin{cor}\label{cor:bow:quadratic}
Under the conditions of Theorem~\ref{thm:bow:quadratic},
suppose each $f_t(\cdot):\X\to\R_+$ with minimizer
$\bv_t$ satisfies the $\lambda$-quadratic growth condition.
Choose $E^0$ and $E^1$ as the algorithm in \eqref{eqn:2}
with $\gamma=\frac{\lambda}{\lambda+\sqrt{\lambda}}$
and $\gamma=1$, respectively.
Then
\begin{equation}\label{eqn:qbow:competitive}
\sum_{t=1}^T
\left(
f_t(\x_t)+\frac{1}{2}\|\x_t-\x_{t-1}\|^2
\right)
\leq
\left(1+\frac{2}{\sqrt{\lambda}}\right)
\sum_{t=1}^T
\left(
f_t(\u_t)+\frac{1}{2}\|\u_t-\u_{t-1}\|^2
\right)
+D^2,
\end{equation}
for every comparator sequence $\u_0,\ldots,\u_T\in\X$
with $\u_0=\x_0$.
Moreover, for every $\u_0,\ldots,\u_T\in\X$,
\begin{equation}\label{eqn:qbow:dynamic}
\begin{split}
&\sum_{t=1}^T
\left(
f_t(\x_t)+\frac{1}{2}\|\x_t-\x_{t-1}\|^2
\right)
-\sum_{t=1}^T
\left(
f_t(\u_t)+\frac{1}{2}\|\u_t-\u_{t-1}\|^2
\right)
\\
\leq{}&
\frac{1}{2}\|\u_0-\x_0\|^2
+DP_T
+8D^2\sqrt{T\log T}
+3D^2
=O\left(P_T+\sqrt{T\log T}\right).
\end{split}
\end{equation}
\end{cor}
\textbf{Remark:} Corollary~\ref{cor:bow:quadratic} preserves the competitive factor
$1+2/\sqrt{\lambda}$ of $E^0$ up to an additive constant $D^2$,
while achieving an $O(P_T+\sqrt{T\log T})$ bound for dynamic regret
with switching cost. Both guarantees hold without the bounded-gradient
Assumption~\ref{ass:1}. When $P_T=\Omega(\sqrt{T\log T})$, the regret
bound matches that of $E^1$ in Lemma~\ref{lem:qbow:dynamic:expert}
up to constant factors.   By the Cauchy--Schwarz inequality, \eqref{eqn:qbow:dynamic} also yields a dynamic-regret
bound of
\[
O\left(
\sqrt{T\sum_{t=1}^T\|\u_t-\u_{t-1}\|^2}
+\sqrt{T\log T}
\right).
\]

\section{Analysis} \label{sec:proof}
In this section, we prove the main results and defer the proofs of supporting lemmas to the appendices. 

\subsection{Proof of Theorem~\ref{thm:ratio:poly}}
Recall that $\x_t$ is the minimizer of $f_t(\cdot)$, which is $\alpha$-polyhedral. When $t \geq 2$, we have
\[
\begin{split}
 & f_t(\x_t) + \|\x_t - \x_{t-1}\|  \\
\leq & f_t(\x_t) + \|\x_t - \u_{t}\|  + \|\u_{t} -\u_{t-1}\|  + \|\u_{t-1} - \x_{t-1}\|\\
\overset{(\ref{eqn:polyhedral})}{\leq} &   f_t(\x_t) + \frac{1}{\alpha} \big( f_t(\u_{t}) - f_t(\x_t) \big)  + \frac{1}{\alpha} \big( f_{t-1}(\u_{t-1}) - f_{t-1}(\x_{t-1}) \big)  + \|\u_{t} -\u_{t-1}\|.
\end{split}
\]
For $t=1$, we have
\[
\begin{split}
  f_1(\x_1) + \|\x_1 - \x_{0}\|  \leq & f_1(\x_1) + \|\x_1 - \u_{1}\|  + \|\u_{1} -\u_{0}\|  + \|\u_{0} - \x_{0}\|\\
=& f_1(\x_1) + \|\x_1 - \u_{1}\|  + \|\u_{1} -\u_{0}\| \\
\overset{(\ref{eqn:polyhedral})}{\leq} &   f_1(\x_1) + \frac{1}{\alpha} \big( f_1(\u_{1}) - f_1(\x_1) \big)  + \|\u_{1} -\u_{0}\|.
\end{split}
\]

Summing over all the iterations, we have
\begin{align}
&\sum_{t=1}^T
  \bigl(f_t(\x_t)+\|\x_t-\x_{t-1}\|\bigr)
  \notag\\
\leq&
  \sum_{t=1}^T f_t(\x_t)
  +\frac{1}{\alpha}\sum_{t=1}^T
    \bigl(f_t(\u_t)-f_t(\x_t)\bigr)
  +\frac{1}{\alpha}\sum_{t=2}^T
    \bigl(f_{t-1}(\u_{t-1})-f_{t-1}(\x_{t-1})\bigr)
  +\sum_{t=1}^T\|\u_t-\u_{t-1}\|
  \notag\displaybreak[1]\\
\leq&
  \sum_{t=1}^T f_t(\x_t)
  +\frac{2}{\alpha}\sum_{t=1}^T
    \bigl(f_t(\u_t)-f_t(\x_t)\bigr)
  +\sum_{t=1}^T\|\u_t-\u_{t-1}\|
  \notag\\
=&
  \frac{2}{\alpha}\sum_{t=1}^T f_t(\u_t)
  +\sum_{t=1}^T\|\u_t-\u_{t-1}\|
  +\sum_{t=1}^T
    \left(1-\frac{2}{\alpha}\right)f_t(\x_t)
  \label{eqn:naive:sum}
\end{align}
where the second inequality follows from the fact that
$f_T(\x_T)\leq f_T(\u_T)$.

Thus, if $\alpha \geq 2$, we have
\begin{equation} \label{eqn:naive:c1}
\begin{split}
  \sum_{t=1}^T  \big( f_t(\x_t) + \|\x_t - \x_{t-1}\| \big)  \overset{(\ref{eqn:naive}), (\ref{eqn:naive:sum})}{\leq} & \frac{2}{\alpha} \sum_{t=1}^T  f_t(\u_{t}) + \sum_{t=1}^T  \|\u_{t} -\u_{t-1}\|  +  \sum_{t=1}^T  \left(\! 1- \frac{2}{\alpha}\!\right) f_t(\u_t) \\
\leq &  \sum_{t=1}^T  \big( f_t(\u_{t}) + \|\u_{t} -\u_{t-1}\|  \big)  \\
\end{split}
\end{equation}
which implies the naive algorithm is $1$-competitive. Otherwise, we have
\begin{equation} \label{eqn:naive:c2}
\begin{split}
 & \sum_{t=1}^T  \big( f_t(\x_t) + \|\x_t - \x_{t-1}\| \big)  \\
\overset{(\ref{eqn:naive:sum})}{\leq}  & \frac{2}{\alpha} \sum_{t=1}^T f_t(\u_{t}) + \sum_{t=1}^T  \|\u_{t} -\u_{t-1}\|  \leq \frac{2}{\alpha} \sum_{t=1}^T  \big( f_t(\u_{t}) + \|\u_{t} -\u_{t-1}\| \big).
\end{split}
\end{equation}
Combining (\ref{eqn:naive:c1}) and (\ref{eqn:naive:c2}) establishes the upper bound.

To prove tightness, consider $\X=[-1,1]$, $x_0=0$, and
$f_t(x)=\alpha|x-v_t|$, where $v_t=(-1)^t$.
The naive algorithm chooses $x_t=v_t$ and incurs zero
hitting cost and total switching cost $2T-1$.
In contrast, the comparator sequence $u_t=0$ for all
$t=0,\ldots,T$ incurs total cost $\alpha T$.
Since the offline optimal cost is at most $\alpha T$,
the ratio between the naive algorithm's cost and the
offline optimal cost is at least
\[
\frac{2T-1}{\alpha T},
\]
which converges to $2/\alpha$ as $T\to\infty$.
Together with the fact that no competitive ratio can
be smaller than $1$, this proves the claimed tightness.

\subsection{Proof of Proposition~\ref{prop:greedy:poly}}

We distinguish three cases according to the value of $\gamma$.

\textbf{Case 1: $\gamma<\alpha$.}
Let $\bv_t$ be the minimizer of $f_t$.
By the polyhedral growth condition and the triangle
inequality, for every $\x\in\X$,
\[
\begin{split}
    f_t(\x)+\gamma\|\x-\x_{t-1}\| \geq & f_t(\bv_t)+ \alpha \|\x-\bv_t\| +\gamma\|\x-\x_{t-1}\| \\
    \geq &    f_t(\bv_t) +(\alpha-\gamma)\|\x-\bv_t\|+\gamma\|\bv_t-\x_{t-1}\|\\
    \geq &f_t(\bv_t) +\gamma\|\bv_t-\x_{t-1}\| ,
\end{split}
\]
where the last inequality is strict whenever $\x\neq\bv_t$, since $\alpha-\gamma>0$. Hence, $\bv_t$ is the unique minimizer of the greedy objective. Thus, the greedy algorithm coincides with the naive algorithm, whose exact worst-case competitive ratio is $\max\left(1,\frac{2}{\alpha}\right)$ by Theorem~\ref{thm:ratio:poly}.

\textbf{Case 2: $\gamma>\alpha$.}
Consider $\X=[0,1]$, $x_0=0$, and
$f_t(x)=\alpha(1-x)$ for every $t$.
Each hitting cost is nonnegative and $\alpha$-polyhedral,
with minimizer $1$.
Whenever $x_{t-1}=0$, the greedy objective becomes
\[
f_t(x)+\gamma|x-x_{t-1}|
=\alpha+(\gamma-\alpha)x,
\]
which is uniquely minimized at $0$.
Consequently, the algorithm remains at $0$ and incurs total
cost $\alpha T$.
The comparator with $u_0=0$ and $u_t=1$ for $t\geq1$
incurs total cost $1$.
The competitive ratio is therefore unbounded as $T\to\infty$.

\textbf{Case 3: $\gamma=\alpha$.}
For any $\varepsilon>0$, consider $\X=[-1,1]$, $x_0=0$,
and
\[
f_t(x)=(\alpha+\varepsilon)|x-v_t|, \  v_t=(-1)^t.
\]
These functions satisfy the $\alpha$-polyhedral condition.
Since $\gamma=\alpha<\alpha+\varepsilon$, the argument
in the first case implies that the greedy algorithm
uniquely selects $x_t=v_t$.
Its total cost is $2T-1$, whereas the comparator
$u_t=0$ for all $t=0,\ldots,T$ incurs total cost
$(\alpha+\varepsilon)T$.
Hence, the competitive ratio is at least
\[
\frac{2T-1}{(\alpha+\varepsilon)T}.
\]
Letting $T\to\infty$ and then $\varepsilon\downarrow0$ gives a lower bound of $2/\alpha$.
Together with the universal lower bound of $1$, this establishes the claimed bound in the boundary case.
The greedy decisions in this construction are unique, so the lower bound holds regardless of how ties are broken.

\subsection{Proof of Theorem~\ref{thm:poly:lower}}
The upper bound follows from Theorem~\ref{thm:ratio:poly}.
Since every competitive ratio is at least $1$, it remains
to establish the lower bound $2/\alpha$ for
$1\leq\alpha<2$.

Fix any deterministic online algorithm.
Consider $\X=[0,1]$, initial point $x_0=0$, and horizon $T=2$.
In the first round, let
\[
f_1(x)=\alpha(1-x)+(\alpha+2)\min\{x,1-x\}.
\]
This function is nonnegative, has unique minimizer $1$,
and satisfies
\[
f_1(x)-f_1(1)\geq\alpha|x-1|, \  \forall x\in[0,1].
\]
Thus, $f_1$ is $\alpha$-polyhedral.
Write $x=x_1$ for the algorithm's first decision and
$y=x_2$ for its second decision.
We distinguish two cases.

\textbf{Case 1: $x\leq1/2$.}
Let $f_2(y)=\alpha(1-y)$.
Since $\alpha\geq1$, the second-round hitting and
switching costs satisfy
\[
f_2(y)+|y-x|
\geq 1-y+|y-x|
\geq 1-x.
\]
Therefore, the algorithm's total cost is at least
\[
f_1(x)+x+(1-x)
=f_1(x)+1
=\alpha+2x+1
\geq\alpha+1.
\]
In contrast, the comparator sequence with $u_0=0$
and $u_1=u_2=1$ incurs total cost $1$.
Hence, the ratio between the algorithm's cost and the
offline optimal cost is at least
\[
\alpha+1\geq\frac{2}{\alpha},
\]
where the last inequality follows from $\alpha\geq1$.

\textbf{Case 2: $x>1/2$.}
Let $f_2(y)=\alpha y$.
Again, since $\alpha\geq1$, the second-round hitting
and switching costs satisfy
\[
f_2(y)+|y-x|
\geq y+|y-x|
\geq x.
\]
Therefore, the algorithm's total cost is at least
\[
f_1(x)+2x
=2+2\alpha(1-x)
\geq2.
\]
The comparator sequence with $u_0=u_1=u_2=0$
incurs total cost $\alpha$.
Thus, the ratio between the algorithm's cost and the
offline optimal cost is at least $2/\alpha$.

Both choices of $f_2$ are nonnegative and
$\alpha$-polyhedral.
Although the choice of $f_2$ depends on the first
decision $x$, this decision is determined by $f_1$
for the fixed deterministic algorithm.
Consequently, the construction specifies a fixed
two-round input sequence for that algorithm. Combining this lower bound with the upper bound in
Theorem~\ref{thm:ratio:poly} proves that the naive
algorithm achieves the optimal competitive ratio
$\max\left(1,\frac{2}{\alpha}\right)$ for $\alpha\geq1$.

\subsection{Proof of Theorem~\ref{thm:naive:quadratic}}
We will make use of the following basic inequality for the squared $\ell_2$-norm \citep[Lemma 12]{NIPS2019_8463}.
\begin{equation}\label{eqn:inequality:squared}
\|\x+\y\|^2 \leq (1+\rho) \|\x\|^2 + \left( 1+ \frac{1}{\rho}\right) \|\y\|^2, \ \forall \rho > 0.
\end{equation}

When $t \geq 2$, we have
\[
\begin{split}
 & f_t(\x_t) + \frac{1}{2}\|\x_t - \x_{t-1}\|^2  \\
\overset{(\ref{eqn:inequality:squared})}{\leq} & f_t(\x_t) +  \frac{1+\rho}{2} \|\u_{t} -\u_{t-1}\|^2 + \frac{1}{2}\left( 1+ \frac{1}{\rho}\right) \|\x_t - \x_{t-1} -\u_{t} +\u_{t-1}\|^2 \\
\overset{(\ref{eqn:inequality:squared})}{\leq} & f_t(\x_t) + \frac{1+\rho}{2}  \|\u_{t} -\u_{t-1}\|^2 + \left( 1+ \frac{1}{\rho}\right)  \big( \|\u_{t} -\x_t \|^2 + \|\u_{t-1} -\x_{t-1}\|^2 \big) \\
\overset{(\ref{eqn:quadratic:growth})}{\leq} & f_t(\x_t) + \frac{1+\rho}{2}  \|\u_{t} -\u_{t-1}\|^2 + \frac{2}{\lambda} \left( 1+ \frac{1}{\rho}\right)  \big( f_t(\u_{t})- f_t(\x_t)  + f_{t-1}(\u_{t-1})-f_{t-1}(\x_{t-1}) \big) .
\end{split}
\]
For $t=1$, we have
\[
 f_1(\x_1) + \frac{1}{2} \|\x_1 - \x_{0}\|^2  \overset{(\ref{eqn:inequality:squared}),(\ref{eqn:quadratic:growth}) }{\leq}    f_1(\x_1) +\frac{1+\rho}{2}  \|\u_{1} -\u_{0}\|^2 + \frac{2}{\lambda} \left( 1+ \frac{1}{\rho}\right)  \big( f_1(\u_{1})- f_1(\x_1)   \big).
\]

Summing over all the iterations, we have
\begin{equation} \label{eqn:naive:sum:quadratic}
\begin{split}
 & \sum_{t=1}^T  \left( f_t(\x_t) + \frac{1}{2} \|\x_t - \x_{t-1}\|^2 \right)  \\
\leq & \sum_{t=1}^T f_t(\x_t)+\frac{1+\rho}{2}   \sum_{t=1}^T  \|\u_{t} -\u_{t-1}\|^2  + \frac{2}{\lambda} \left( 1+ \frac{1}{\rho}\right)  \sum_{t=1}^T  \big( f_t(\u_{t}) - f_t(\x_t) \big) \\
& + \frac{2}{\lambda} \left( 1+ \frac{1}{\rho}\right)  \sum_{t=2}^T  \big( f_{t-1}(\u_{t-1}) - f_{t-1}(\x_{t-1}) \big)   \\
\leq & \sum_{t=1}^T f_t(\x_t)+\frac{1+\rho}{2}   \sum_{t=1}^T  \|\u_{t} -\u_{t-1}\|^2  + \frac{4}{\lambda} \left( 1+ \frac{1}{\rho}\right)  \sum_{t=1}^T  \big( f_t(\u_{t}) - f_t(\x_t) \big) \\
=&   \frac{4}{\lambda} \left( 1+ \frac{1}{\rho}\right) \sum_{t=1}^T  f_t(\u_{t}) + \frac{1+\rho}{2}   \sum_{t=1}^T  \|\u_{t} -\u_{t-1}\|^2 + \left( 1- \frac{4}{\lambda} \left( 1+ \frac{1}{\rho}\right) \right) \sum_{t=1}^T f_t(\x_t).
\end{split}
\end{equation}

First, we consider the case that
\begin{equation} \label{eqn:quadratic:rho}
 1- \frac{4}{\lambda} \left( 1+ \frac{1}{\rho}\right)  \leq 0  \Leftrightarrow    \frac{\lambda }{4} \leq 1 +\frac{1}{\rho}
\end{equation}
and have
\[
\begin{split}
&\sum_{t=1}^T  \left( f_t(\x_t) + \frac{1}{2} \|\x_t - \x_{t-1}\|^2 \right) \\
 \overset{(\ref{eqn:naive:sum:quadratic}), (\ref{eqn:quadratic:rho})}{\leq}    & \frac{4}{\lambda} \left( 1+ \frac{1}{\rho}\right) \sum_{t=1}^T  f_t(\u_{t}) + \frac{1+\rho}{2}   \sum_{t=1}^T  \|\u_{t} -\u_{t-1}\|^2\\
\leq & \max\left( \frac{4}{\lambda} \left(1+ \frac{1}{\rho} \right), 1 + \rho\right) \sum_{t=1}^T \left( f_t(\u_{t}) + \frac{1}{2} \|\u_{t} -\u_{t-1}\|^2 \right).
\end{split}
\]
To minimize the competitive ratio, we set
\[
\frac{4}{\lambda} \left(1+ \frac{1}{\rho} \right) = 1 + \rho \Rightarrow \rho=\frac{4}{\lambda}
\]
and obtain
\begin{equation} \label{eqn:ratio:quadratic:navie:1}
\begin{split}
\sum_{t=1}^T  \left( f_t(\x_t) + \frac{1}{2} \|\x_t - \x_{t-1}\|^2 \right)  \leq \left( 1 + \frac{4}{\lambda} \right) \sum_{t=1}^T \left( f_t(\u_{t}) + \frac{1}{2} \|\u_{t} -\u_{t-1}\|^2 \right).
\end{split}
\end{equation}

Next, we study the case that
\[
 1- \frac{4}{\lambda} \left( 1+ \frac{1}{\rho}\right)  \geq 0  \Leftrightarrow    \frac{\lambda }{4} \geq 1 +\frac{1}{\rho}
\]
which only happens when $\lambda > 4$. Then, we have
\[
 \sum_{t=1}^T  \left( f_t(\x_t) + \frac{1}{2} \|\x_t - \x_{t-1}\|^2 \right)  \overset{(\ref{eqn:naive}), (\ref{eqn:naive:sum:quadratic})}{\leq}     \sum_{t=1}^T  f_t(\u_{t}) + \frac{1+\rho}{2}   \sum_{t=1}^T  \|\u_{t} -\u_{t-1}\|^2.
\]
To minimize the competitive ratio, we set $\rho  = \frac{4}{\lambda-4}$, and obtain
\[
\sum_{t=1}^T  \left( f_t(\x_t) + \frac{1}{2} \|\x_t - \x_{t-1}\|^2 \right)  \leq  \frac{\lambda}{\lambda -4 }   \sum_{t=1}^T \left( f_t(\u_{t}) + \frac{1}{2}     \|\u_{t} -\u_{t-1}\|^2 \right)
\]
which is worse than (\ref{eqn:ratio:quadratic:navie:1}). So, we keep (\ref{eqn:ratio:quadratic:navie:1}) as the final result.

\subsection{Proof of Theorem~\ref{thm:greedy:convex:quadratic}}
Since $f_t(\cdot)$ is convex, the objective function of (\ref{eqn:2}) is $\gamma$-strongly convex. From the quadratic growth property of strongly convex functions \citep{COLT:Hazan:2011}, we have
\begin{equation}\label{eqn:opt:convex:quadratic}
f_t(\x_t) + \frac{\gamma}{2}\|\x_t-\x_{t-1}\|^2 + \frac{\gamma}{2} \|\u-\x_t\|^2 \leq f_t(\u) + \frac{\gamma}{2}\|\u-\x_{t-1}\|^2, \ \forall \u \in \X.
\end{equation}

Similar to previous studies \citep{bansal_et_al:LIPIcs}, the analysis uses an amortized local competitiveness argument, using the potential function $c \|\x_t -\u_t\|^2$. We proceed to bound $f_t(\x_t) + \frac{1}{2} \|\x_t-\x_{t-1}\|^2 + c \|\x_t -\u_t\|^2 - c\|\x_{t-1}-\u_{t-1}\|^2$, and have
\[
\begin{split}
&  f_t(\x_t) + \frac{1}{2}\|\x_t-\x_{t-1}\|^2 + c \|\x_t -\u_t\|^2 - c\|\x_{t-1}-\u_{t-1}\|^2\\
\overset{(\ref{eqn:inequality:squared})}{\leq}  &  f_t(\x_t) + \frac{1}{2}\|\x_t-\x_{t-1}\|^2+ c \big( 2 \|\x_t - \bv_t\|^2 + 2\|\bv_t-\u_t\|^2 \big)- c\|\x_{t-1}-\u_{t-1}\|^2\\
\overset{(\ref{eqn:quadratic:growth})}{\leq}  & \left(1+\frac{4c }{\lambda} \right) f_t(\x_t) + \frac{1}{2} \|\x_t-\x_{t-1}\|^2 + \frac{4c}{\lambda} f_t (\u_t) - c\|\x_{t-1}-\u_{t-1}\|^2\\
= & \left(1 +\frac{4c }{\lambda} \right) \left( f_t(\x_t) + \frac{\lambda}{2(\lambda + 4c) } \|\x_t-\x_{t-1}\|^2 \right) + \frac{4c}{\lambda} f_t (\u_t) - c\|\x_{t-1}-\u_{t-1}\|^2.
\end{split}
\]

Under the condition
\begin{equation} \label{eqn:convex:quadratic:constraint:1}
\frac{\lambda}{\lambda + 4c} \leq \gamma,
\end{equation}
we have
\[
\begin{split}
&  f_t(\x_t) + \frac{1}{2}\|\x_t-\x_{t-1}\|^2 + c \|\x_t -\u_t\|^2 - c\|\x_{t-1}-\u_{t-1}\|^2\\
\leq & \left(1+\frac{4c }{\lambda} \right) \left( f_t(\x_t) + \frac{\gamma}{2} \|\x_t-\x_{t-1}\|^2 \right)  + \frac{4c}{\lambda} f_t (\u_t)  - c\|\x_{t-1}-\u_{t-1}\|^2\\
\overset{(\ref{eqn:opt:convex:quadratic})}{\leq} & \left(1+\frac{4c }{\lambda}\right) \left( f_t(\u_t) + \frac{\gamma}{2} \|\u_t-\x_{t-1}\|^2 -\frac{\gamma}{2} \|\u_t-\x_t\|^2 \right)  + \frac{4c}{\lambda} f_t (\u_t)  - c\|\x_{t-1}-\u_{t-1}\|^2\\
= & \left(1+\frac{8c }{\lambda}\right) f_t (\u_t) + \frac{\gamma(\lambda + 4c) }{2\lambda} \|\u_t-\x_{t-1}\|^2  -  \frac{\gamma(\lambda + 4c) }{2\lambda} \|\u_t-\x_t\|^2- c\|\x_{t-1}-\u_{t-1}\|^2.
\end{split}
\]

Summing over all iterations and using $\x_0=\u_0$, we obtain
\[
\begin{aligned}
&\sum_{t=1}^T
\left(f_t(\x_t)+\frac12\|\x_t-\x_{t-1}\|^2\right)
+c\|\x_T-\u_T\|^2
\\
\leq&
\left(1+\frac{8c}{\lambda}\right)
\sum_{t=1}^T f_t(\u_t)
+\frac{\gamma(\lambda+4c)}{2\lambda}
\sum_{t=1}^T\|\u_t-\x_{t-1}\|^2-
\left(\frac{\gamma(\lambda+4c)}{2\lambda}+c\right)
\sum_{t=1}^T\|\x_{t-1}-\u_{t-1}\|^2.
\end{aligned}
\]
Here we have discarded a nonpositive terminal term. For any $\rho>0$, \eqref{eqn:inequality:squared} implies
\[
\|\x_{t-1}-\u_{t-1}\|^2
\geq
\frac{1}{1+\rho}\|\x_{t-1}-\u_t\|^2
-\frac{1}{\rho}\|\u_t-\u_{t-1}\|^2.
\]
To make the resulting coefficient of
$\sum_{t=1}^T\|\u_t-\x_{t-1}\|^2$ nonpositive, we impose
\begin{equation}
\label{eqn:convex:quadratic:constraint:2}
\frac{\gamma(\lambda+4c)}{2\lambda}
\leq
\left(\frac{\gamma(\lambda+4c)}{2\lambda}+c\right)
\frac{1}{1+\rho}
\quad\Longleftrightarrow\quad
\frac{\gamma(\lambda+4c)}{2\lambda}
\leq\frac{c}{\rho}.
\end{equation}
Combining these inequalities and dropping the nonnegative
term $c\|\x_T-\u_T\|^2$ gives
\[
\begin{aligned}
&\sum_{t=1}^T
\left(f_t(\x_t)+\frac12\|\x_t-\x_{t-1}\|^2\right)
\\
\leq&
\left(1+\frac{8c}{\lambda}\right)
\sum_{t=1}^T f_t(\u_t)
+
\left(\frac{\gamma(\lambda+4c)}{2\lambda}+c\right)
\frac{1}{\rho}
\sum_{t=1}^T\|\u_t-\u_{t-1}\|^2
\\
\leq& \max\left(
1+\frac{8c}{\lambda},
\left(\frac{\gamma(\lambda+4c)}{2\lambda}+c\right)
\frac{2}{\rho}
\right)
\sum_{t=1}^T
\left(f_t(\u_t)+\frac12\|\u_t-\u_{t-1}\|^2\right).
\end{aligned}
\]

Next, we minimize the competitive ratio under the constraints in (\ref{eqn:convex:quadratic:constraint:1}) and (\ref{eqn:convex:quadratic:constraint:2}), which can be summarized as
\[
\frac{\lambda}{\lambda + 4c} \leq \gamma \leq  \frac{\lambda}{\lambda + 4c}  \frac{2c}{\rho}.
\]
We first set $c=\frac{\rho}{2}$ and $\gamma=\frac{\lambda}{\lambda + 4c}$, and obtain
\[
\sum_{t=1}^T \left( f_t(\x_t) + \frac{1}{2}\|\x_t-\x_{t-1}\|^2 \right)
 \leq  \max\left(1 +\frac{4 \rho}{\lambda},  1 + \frac{1 }{\rho} \right) \sum_{t=1}^T  \left(f_t (\u_t) + \frac{1}{2}\|\u_t - \u_{t-1}\|^2\right).
\]
Then, we set
\[
1 +\frac{4 \rho}{\lambda}= 1 + \frac{1 }{\rho} \Rightarrow  \rho= \frac{\sqrt{\lambda}}{2}.
\]
As a result, the competitive ratio is
\[
1 + \frac{1 }{\rho}  = 1 + \frac{2}{\sqrt{\lambda}},
\]
and the parameter is
\[
 \gamma=\frac{\lambda}{\lambda + 4c}  = \frac{\lambda}{\lambda + 2 \rho} = \frac{\lambda}{\lambda + \sqrt{\lambda}}  .
\]

\subsection{Proof of Theorem~\ref{thm:ader}}
The analysis is similar to the proof of Theorem 3 of \citet{Adaptive:Dynamic:Regret:NIPS}.  By the initialization in Algorithms~\ref{alg:1} and~\ref{alg:2}, the
meta-algorithm and all experts share an arbitrary initial point
$\x_0\in\X$; that is,
\begin{equation} \label{eqn:condit:at:zero}
 \x_0^\eta=\x_0, \ \forall \eta \in \H.
\end{equation}

First, we bound the dynamic regret with switching cost of the meta-algorithm w.r.t.~all experts simultaneously.
\begin{lem} \label{lem:meta} Under Assumptions~\ref{ass:2} and \ref{ass:1}, and setting $\beta= \frac{2}{(2G+1) D} \sqrt{\frac{2 }{5 T}}$, we have
\begin{equation}\label{thm:main:1}
 \sum_{t=1}^T \big( s_t(\x_t) + \|\x_t -\x_{t-1}\|\big) - \sum_{t=1}^T \big(s_t(\x_t^\eta)  + \|\x_t^\eta- \x_{t-1}^\eta\| \big) \leq  (2G+1) D\sqrt{\frac{5  T }{8}} \left( \ln \frac{1}{w_1^\eta} + 1\right)
\end{equation}
for each $\eta \in \H$.
\end{lem}

Next, we bound the dynamic regret with switching cost of each expert w.r.t.~any comparator sequence $\u_0, \u_1,\ldots,\u_T \in \X$.
\begin{lem} \label{lem:expert} Under Assumptions \ref{ass:2} and \ref{ass:1},  we have
\begin{equation}\label{eqn:main:2}
\sum_{t=1}^T \big(s_t(\x_t^\eta)  + \|\x_t^\eta- \x_{t-1}^\eta\| \big) - \sum_{t=1}^T  s_t(\u_t) \leq \frac{D^2}{2 \eta}    + \frac{D}{\eta} \sum_{t=1}^T  \|\u_t - \u_{t-1}\| + \eta T \left( \frac{G^2}{2}  +G\right).
\end{equation}
\end{lem}

Then, we show that for any sequence of  comparators $\u_0, \u_1,\ldots,\u_T \in \X$ there exists an $\eta_k \in \H$ such that the RHS of (\ref{eqn:main:2}) is almost minimal. If we minimize the RHS of (\ref{eqn:main:2}) exactly, the optimal step size is
\begin{equation}\label{thm:main:3}
\eta^*(P_T)= \sqrt{\frac{D^2+ 2 D P_T}{T (G^2+2G)}}.
\end{equation}
From Assumption~\ref{ass:2}, we have the following bound of the path-length
\begin{equation}\label{thm:main:9}
0 \leq P_T= \sum_{t=1}^T  \|\u_t - \u_{t-1}\| \overset{(\ref{eqn:domain})}{\leq} TD.
\end{equation}
Thus
\[
\sqrt{\frac{D^2}{T (G^2+2G)}}\leq \eta^*(P_T) \leq \sqrt{\frac{D^2+2TD^2}{T (G^2+2G)}}.
\]
From our construction of $\H$ in (\ref{eqn:def:H}), it is easy to verify that
\[
\min \H= \sqrt{\frac{D^2}{T (G^2+2G)}}, \textrm{ and  } \max \H \geq \sqrt{\frac{D^2+2TD^2}{T (G^2+2G)}}.
\]
As a result, for any possible value of $P_T$, there exists a step size $\eta_k \in \H$ with $k$ defined in (\ref{eqn:def:k}), such that
\begin{equation}\label{thm:main:4}
\eta_k = 2^{k-1}  \sqrt{\frac{D^2}{T (G^2+2G)}}  \leq \eta^*(P_T) \leq 2 \eta_k.
\end{equation}

Substituting $\eta_k$ into (\ref{eqn:main:2}), we obtain the following bound for expert $E^{\eta_k}$:
\begin{equation}\label{thm:main:5}
\begin{split}
& \sum_{t=1}^T \big(s_t(\x_t^{\eta_k})  + \|\x_t^{\eta_k}- \x_{t-1}^{\eta_k}\| \big) - \sum_{t=1}^T  s_t(\u_t) \\
\leq &\frac{D^2}{2 \eta_k}    + \frac{D}{{\eta_k}} \sum_{t=1}^T  \|\u_t - \u_{t-1}\| + \eta_k T \left( \frac{G^2}{2}  +G\right)\\
\overset{(\ref{thm:main:4})}{\leq} & \frac{D^2}{\eta^*(P_T)}    + \frac{2D}{{\eta^*(P_T)}} \sum_{t=1}^T  \|\u_t - \u_{t-1}\| + \eta^*(P_T) T \left( \frac{G^2}{2}  +G\right) \\
 \overset{(\ref{thm:main:3})}{=} & \frac{3}{2}\sqrt{T (G^2+2G)(D^2+ 2 D P_T)}.
\end{split}
\end{equation}

From (\ref{eqn:initial:weight}), we know the initial weight of expert $E^{\eta_k}$ is
\[
w_1^{\eta_k}=\frac{C}{k(k+1)} \geq \frac{1}{k(k+1)} \geq \frac{1}{(k+1)^2}.
\]
Combining with (\ref{thm:main:1}), we obtain the relative performance of the meta-algorithm w.r.t.~expert $E^{\eta_k}$:
\begin{equation}\label{thm:main:6}
 \sum_{t=1}^T \big( s_t(\x_t) + \|\x_t -\x_{t-1}\|\big) - \sum_{t=1}^T \big(s_t(\x_t^{\eta_k})  + \|\x_t^{\eta_k}- \x_{t-1}^{\eta_k}\| \big) \leq  (2G+1) D\sqrt{\frac{5  T }{8}} \left[1+ 2\ln (k+1)\right].
\end{equation}

From  (\ref{thm:main:5}) and (\ref{thm:main:6}), we derive the following upper bound for dynamic regret with switching cost
\begin{equation}\label{thm:main:7}
\begin{split}
& \sum_{t=1}^T \big( s_t(\x_t) + \|\x_t -\x_{t-1}\|\big)-  \sum_{t=1}^T  s_t(\u_t) \\
\leq & \frac{3}{2}\sqrt{T (G^2+2G)(D^2+ 2 D P_T)}+ (2G+1) D\sqrt{\frac{5  T }{8}} \left[1+ 2\ln (k+1)\right].
\end{split}
\end{equation}
Finally, from Assumption~\ref{ass:0}, we have
\begin{equation}\label{thm:main:8}
f_t(\x_t) - f_t(\u_t) \leq \langle \nabla f_t(\x_t), \x_t-\u_t \rangle  \overset{(\ref{eqn:surrogate})}{=}   s_t(\x_t) - s_t(\u_t).
\end{equation}
We complete the proof by combining (\ref{thm:main:7}) and (\ref{thm:main:8}).

\subsection{Proof of Theorem~\ref{thm:lookahead:ader}}
The analysis is similar to that of Theorem~\ref{thm:ader}. The difference is that we need to take into account the lookahead property of the meta-algorithm and the expert-algorithm.

First, we bound the dynamic regret with switching cost of the meta-algorithm w.r.t.~all experts simultaneously.
\begin{lem} \label{lem:lookhead:meta} Under Assumptions~\ref{ass:0} and~\ref{ass:2}, and setting $\beta= \frac{1}{D}  \sqrt{\frac{2}{T}}$, we have
\begin{equation}\label{eqn:lookhead:2}
 \sum_{t=1}^T \big( f_t(\x_t) + \|\x_t -\x_{t-1}\|\big) - \sum_{t=1}^T \big(f_t(\x_t^\eta)  + \|\x_t^\eta- \x_{t-1}^\eta\| \big) \leq  D \sqrt{ \frac{T}{2}}  \left( \ln \frac{1}{w_0^\eta} +1 \right)
\end{equation}
for each $\eta \in \H$.
\end{lem}

Next, we bound the dynamic regret with switching cost of each expert w.r.t.~any comparator sequence $\u_0, \u_1,\ldots,\u_T \in \X$.
\begin{lem} \label{lem:lookhead:expert} Under Assumptions~\ref{ass:0} and \ref{ass:2},  we have
\begin{equation}\label{eqn:lookhead:3}
\sum_{t=1}^T \big(f_t(\x_t^\eta)  + \|\x_t^\eta- \x_{t-1}^\eta\| \big) - \sum_{t=1}^T  f_t(\u_t) \leq \frac{D^2}{2 \eta}    + \frac{D}{\eta} \sum_{t=1}^T \|\u_t - \u_{t-1}\|  + \frac{\eta T}{2}.
\end{equation}
\end{lem}

The rest of the proof is almost identical to that of Theorem~\ref{thm:ader}. We will show that  for any sequence of  comparators $\u_0, \u_1,\ldots,\u_T \in \X$ there exists an $\eta_k \in \H$ such that the RHS of (\ref{eqn:lookhead:3}) is almost minimal. If we minimize the RHS of (\ref{eqn:lookhead:3}) exactly, the optimal step size is
\begin{equation}\label{eqn:lookhead:4}
\eta^*(P_T)= \sqrt{\frac{D^2+ 2 D P_T}{T}}.
\end{equation}
From (\ref{thm:main:9}), we know that
\[
\sqrt{\frac{D^2}{T}}\leq \eta^*(P_T) \leq \sqrt{\frac{D^2+2TD^2}{T}}.
\]
From our construction of $\H$ in (\ref{eqn:def:H:New}), it is easy to verify that
\[
\min \H= \sqrt{\frac{D^2}{T }}, \textrm{ and  } \max \H \geq \sqrt{\frac{D^2+2TD^2}{T}}.
\]
As a result, for any possible value of $P_T$, there exists a step size $\eta_k \in \H$ with $k$ defined in (\ref{eqn:def:k}), such that
\begin{equation}\label{eqn:lookhead:5}
\eta_k = 2^{k-1}  \sqrt{\frac{D^2}{T}}  \leq \eta^*(P_T) \leq 2 \eta_k.
\end{equation}

Plugging $\eta_k$ into (\ref{eqn:lookhead:3}), we have the following bound for expert $E^{\eta_k}$:
\begin{equation}\label{eqn:lookhead:6}
\begin{split}
& \sum_{t=1}^T \big(f_t(\x_t^{\eta_k})  + \|\x_t^{\eta_k}- \x_{t-1}^{\eta_k}\| \big) - \sum_{t=1}^T  f_t(\u_t) \\
\leq &\frac{D^2}{2 \eta_k}    + \frac{D}{{\eta_k}} \sum_{t=1}^T  \|\u_t - \u_{t-1}\| + \frac{\eta_k T}{2}\\
\overset{(\ref{eqn:lookhead:5})}{\leq} & \frac{D^2}{\eta^*(P_T)} + \frac{2D}{{\eta^*(P_T)}} \sum_{t=1}^T  \|\u_t - \u_{t-1}\| + \frac{\eta^*(P_T) T}{2}\\
 \overset{(\ref{eqn:lookhead:4})}{=} & \frac{3}{2}\sqrt{T (D^2+ 2 D P_T)}.
\end{split}
\end{equation}

From Step 2 of Algorithm~\ref{alg:3}, we know the initial weight of expert $E^{\eta_k}$ is
\[
w_0^{\eta_k}=\frac{C}{k(k+1)} \geq \frac{1}{k(k+1)} \geq \frac{1}{(k+1)^2}.
\]
Combining with (\ref{eqn:lookhead:2}), we obtain the relative performance of the meta-algorithm w.r.t.~expert $E^{\eta_k}$:
\begin{equation}\label{eqn:lookhead:7}
 \sum_{t=1}^T \big( f_t(\x_t) + \|\x_t -\x_{t-1}\|\big) - \sum_{t=1}^T \big(f_t(\x_t^{\eta_k})  + \|\x_t^{\eta_k}- \x_{t-1}^{\eta_k}\| \big) \leq D \sqrt{ \frac{T}{2}}  \left[1+ 2\ln (k+1)\right].
\end{equation}
We complete the proof by summing (\ref{eqn:lookhead:6}) and (\ref{eqn:lookhead:7}) together.

\subsection{Proof of Theorem~\ref{thm:lower}}
The proof is built upon a lower bound of competitive ratio \citep{pmlr-v125-argue20a}. By setting $\gamma=\frac{D}{2 \sqrt{d}}$ in Lemma 12 of \citet{pmlr-v125-argue20a}, we can guarantee that Assumption~\ref{ass:2} is satisfied.  Then, we choose $\mu=0$, $\lambda=1/\gamma$ in that lemma, and obtain the conclusion below.

\begin{lem} \label{thm:lower:ratio} For any deterministic online algorithm $A$ and any fixed value of $d$, there exists a sequence of convex functions $f_1(\cdot), \ldots, f_d(\cdot)$ over the domain $[-\frac{D}{2 \sqrt{d}},\frac{D}{2 \sqrt{d}}]^d$ in the lookahead setting such that
\begin{compactenum}
  \item the sum of the hitting cost and the switching cost of $A$ is at least $\frac{3 \gamma d}{4}=\frac{3 D \sqrt{d}}{8}$;
  \item there exists a fixed point $\u$ whose hitting cost is $0$.
\end{compactenum}
\end{lem}
Although the proof of \citet[Lemma 12]{pmlr-v125-argue20a} uses random signs, an averaging argument guarantees a fixed realization for which the same lower bound holds.

We consider two cases. First, suppose that $0\leq\tau<64D$. Applying Lemma~\ref{thm:lower:ratio} with $d=T$ gives a sequence of convex functions and a fixed point $\u$ whose hitting cost is zero, whereas the online algorithm incurs cost at least $3D\sqrt{T}/8$. Set $\u_0=\u_1=\cdots=\u_T=\u$. This comparator has zero total cost, and hence
the dynamic regret with switching cost is at least 
\[
\frac{3D\sqrt{T}}{8}
\geq \frac{3}{8\sqrt{65}}\sqrt{T(D^2+D\tau)},
\]
where the inequality follows from $\tau<64D$.

Next, suppose that $\tau\geq64D$. Set
\[
K=\left\lfloor\frac{\tau}{64D}\right\rfloor \text{ and } d=\left\lfloor\frac{T}{K}\right\rfloor.
\]
Because $\tau\leq TD$, we have 
\[
1\leq K\leq\frac{T}{64}, \  K\geq\frac{\tau}{128D}, \text {and } d\geq\frac{T}{2K},
\]
where we use $\lfloor x\rfloor\geq x/2$ for every $x\geq1$. Partition the first $Kd$ rounds into $K$ consecutive stages of $d$ rounds each. All stages use the common domain $[-D/(2\sqrt{d}),D/(2\sqrt{d})]^d$, whose Euclidean diameter is $D$. Given the history before each stage, Lemma~\ref{thm:lower:ratio} applies to the continuation
of the online algorithm. Thus, in each stage, there exists a sequence of $d$ convex hitting costs such that the online algorithm incurs cost at least $3D\sqrt{d}/8$, while some fixed point has zero hitting cost throughout that stage. Concatenating
these stagewise constructions gives a total online cost of at least
\[
\frac{3D}{8}K\sqrt{d}.
\]
If $Kd<T$, set the hitting costs in the remaining $T-Kd$ rounds to zero; these additional rounds cannot decrease the cost of the online algorithm.

Construct the comparator by remaining at the fixed point supplied by the lemma during each stage. Set $\u_0$ equal to the fixed point in the first stage, and keep the comparator at its last-stage point during the zero-cost padding rounds. The comparator has zero hitting cost and changes its decision at most $K-1$ times. Since the diameter of the domain is $D$, its path length satisfies
\[
P_T\leq D(K-1)\leq DK\leq\frac{\tau}{64}\leq\tau.
\]
Consequently, the dynamic regret with switching cost is at least
\[
\begin{split}
\frac{3D}{8}K\sqrt{d}-D(K-1)
&\geq \frac{3D}{8}K\sqrt{\frac{T}{2K}}-DK =D\sqrt{TK}\left(\frac{3}{8\sqrt{2}}-\sqrt{\frac{K}{T}}\right)\\
&\geq\frac{D}{8}\sqrt{TK} =\Omega\big(\sqrt{TD\tau}\big)=\Omega\big(\sqrt{T(D^2+D\tau)}\big),
\end{split}
\]
where the first inequality uses $d\geq T/(2K)$; the second inequality follows from $K\leq T/64$; and the last two relations use $K\geq\tau/(128D)$ and $\tau\geq D$, respectively.

\subsection{Proof of Theorem~\ref{thm:bow}}\label{app:proof:bow}
By Lemma~\ref{lem:bow:interpolation}, it suffices to bound the two residual terms in \eqref{eq:bow:interpolation:E0}
and \eqref{eq:bow:interpolation:E1}. To this end, we set $q_0=0$ and define
\begin{align*}
	\Rt_0(\bv;\Gamma;K) &=\sum_{t=1}^T\left(\frac{K}{2}|v_t-v_{t-1}|-v_t\Gamma_t \right),
	\\
	\Rt_1(\bv;\Gamma;K) &=\sum_{t=1}^T\left(\frac{K}{2}|v_t-v_{t-1}|+(1-v_t)\Gamma_t\right).
\end{align*}
Here $\bv=(v_0,\dots,v_T)$ is any sequence of weights in $[0,1]$, $\Gamma=(\Gamma_1,\dots,\Gamma_T)$ is any real-valued gap sequence, and $K>0$ controls the variation penalty. Therefore, \eqref{eq:bow:interpolation:E0} and
\eqref{eq:bow:interpolation:E1} can be written as
\begin{align}
	\sum_{t=1}^T\left(f_t(\x_t)+\|\x_t-\x_{t-1}\|-C_t^0\right)
	&\le
	\Rt_0(\p;\Delta;2D),
	\label{eqn:bow:regret:0}
	\\
	\sum_{t=1}^T\left(f_t(\x_t)+\|\x_t-\x_{t-1}\|-C_t^1\right)
	&\le
	\Rt_1(\p;\Delta;2D).
	\label{eqn:bow:regret:1}
\end{align}
The two terms $\Rt_0(\p;\Delta;2D)$ and $\Rt_1(\p;\Delta;2D)$ are the quantities that we will bound in the following analysis.

To apply the DNP guarantee, we introduce the reference residuals $\Rt_i(\q;\dtb;2D)$, $i\in\{0,1\}$, defined using the reference
weights $q_t$ and clipped gaps $\dtb_t$. Comparing these quantities with the original residuals requires
accounting for the additional variation introduced when UDNP overrides $q_t$. The following lemma shows that this additional penalty is offset by the gain from selecting the expert with the smaller current cost,
so that each original residual is bounded by its reference counterpart. We then apply the DNP guarantee to bound the reference residuals.
\begin{lem}\label{lem:bow:transfer}Fix $K>0$.
Let $\Delta_1,\ldots,\Delta_T\in\R$ and
$q_1,\ldots,q_T\in[0,1]$ be arbitrary sequences,
and set $p_0=q_0=0$.
For $t=1,\ldots,T$, let
$\dtb_t=\Pi_{[-K,K]}[\Delta_t]$ and define
\[
	p_t=\begin{cases}
		1 ,& \Delta_t \ge K \\
		0 ,& \Delta_t \le -K \\
		q_t ,& |\Delta_t| < K
	\end{cases}.
\]
Then
\begin{equation}\label{eqn:bow:transfer}
\Rt_i(\p;\Delta;K)
\le
\Rt_i(\p;\dtb;K)
\le
\Rt_i(\q;\dtb;K),  \  i\in\{0,1\}.
\end{equation}
\end{lem}
The proof proceeds in two steps. For the first inequality, whenever clipping occurs, UDNP selects
the expert with the smaller current cost.
The discarded part of the gap therefore contributes nonpositively
to both residuals. For the second inequality, under the clipped gaps, replacing $q_t$
by $p_t$ reduces the gap-dependent term of each residual by
$K|p_t-q_t|$. Meanwhile, the triangle inequality and $p_0=q_0$ imply that the total increase in the variation penalty is at most
$K\sum_{t=1}^T|p_t-q_t|$.
Thus, the reduction in the gap-dependent terms compensates for
the additional variation penalty.
The full proof is given in Appendix~\ref{app:bow:transfer}.

It remains to bound the reference residuals. The following lemma follows from Theorem~5 of
\citet{Smooth:DNP:NeurIPS}.
\begin{lem}\label{lem:bow:clip_bound}
Fix $K>0$ and let $\dtb_t=\Pi_{[-K,K]}[\Delta_t]$.
Suppose Algorithm~\ref{alg:dnp} runs with $Z=1/T$ and receives $b_t=\dtb_t/K$ at each round.
If $T\geq\max\{8e,16\log T\}$, then
\[
\Rt_0(\q;\dtb;K)\leq K, \ \textrm{ and } \ \Rt_1(\q;\dtb;K)\leq 8K\sqrt{T\log T}+3K.
\]
\end{lem}

By Lemma~\ref{lem:bow:transfer} and Lemma~\ref{lem:bow:clip_bound}, the two residual terms satisfy
\begin{equation}\label{eqn:bow:final:meta}
	\Rt_0(\p;\Delta;2D) \le 2D,  \ \textrm{ and } \ \Rt_1(\p;\Delta;2D)\le  16D\sqrt{T\log T}+6D.
\end{equation}
Combining \eqref{eqn:bow:regret:0} and \eqref{eqn:bow:regret:1} with \eqref{eqn:bow:final:meta} proves
\eqref{eqn:bow:aggregation} and completes the proof.

\subsection{Proof of Corollary~\ref{cor:bow}}
For the competitive-ratio guarantee, applying
Theorem~\ref{thm:ratio:poly} to $E^0$ gives
\begin{align*}
\sum_{t=1}^T\big(f_t(\x_t)+\|\x_t-\x_{t-1}\|\big)
\overset{\eqref{eqn:bow:aggregation}}{\le}&
\sum_{t=1}^T\left( f_t(\x_t^0) + \|\x_t^0- \x_{t-1}^0\| \right)+ 2D \\
\overset{\eqref{eqn:ratio:poly}}{\le} &
\max\left( 1, \frac{2}{\alpha} \right) \sum_{t=1}^T  \left(f_t (\u_t) + \|\u_t - \u_{t-1}\|\right) + 2D
\end{align*}
for every comparator sequence $\u_0,\ldots,\u_T\in\X$ with $\u_0=\x_0$.  For the dynamic-regret guarantee, applying
Theorem~\ref{thm:lookahead:ader} to $E^1$ gives
\begin{align*}
&\sum_{t=1}^T\big(f_t(\x_t)+\|\x_t-\x_{t-1}\|\big
)-\sum_{t=1}^T  f_t(\u_t)
\\
\overset{\eqref{eqn:bow:aggregation}}{\le}&\sum_{t=1}^T \left( f_t(\x_t^1) + \|\x_t^1- \x_{t-1}^1\| \right)-\sum_{t=1}^T  f_t(\u_t) + 16D\sqrt{T\log T} + 6D
\\
\overset{\eqref{eqn:look:sader}}{\le}
&\frac{3}{2}\sqrt{T (D^2+ 2 D \sum_{t=1}^T \|\u_t - \u_{t-1}\|)}+ D \sqrt{ \frac{T}{2}}  \left[1+ 2\ln (k+1)\right] + 16D\sqrt{T\log T} + 6D
. \end{align*}
This completes the proof.

\subsection{Proof of Theorem~\ref{thm:bow:quadratic}}
By Lemma~\ref{lem:qbow:interpolation}, it suffices to bound
the two residual terms in \eqref{eq:qbow:interpolation:E0}
and \eqref{eq:qbow:interpolation:E1}.
Using the notation introduced in the proof of
Theorem~\ref{thm:bow} and setting $q_0=0$, we have
\begin{align}
    \sum_{t=1}^T
    \left(
        f_t(\x_t)
        +\frac12\|\x_t-\x_{t-1}\|^2
        -C_t^0
    \right)
    &\leq \Rt_0(\p;\Delta;D^2),
    \label{eqn:bow:squared_regret:0}
    \\
    \sum_{t=1}^T
    \left(
        f_t(\x_t)
        +\frac12\|\x_t-\x_{t-1}\|^2
        -C_t^1
    \right)
    &\leq \Rt_1(\p;\Delta;D^2).
    \label{eqn:bow:squared_regret:1}
\end{align}
Applying Lemma~\ref{lem:bow:transfer} and
Lemma~\ref{lem:bow:clip_bound} with $K=D^2$ gives
\begin{equation}\label{eqn:qbow:final:meta}
    \Rt_0(\p;\Delta;D^2)\leq D^2, \  \textrm{ and } \   \Rt_1(\p;\Delta;D^2)
    \leq 8D^2\sqrt{T\log T}+3D^2.
\end{equation}

Combining \eqref{eqn:bow:squared_regret:0} and \eqref{eqn:bow:squared_regret:1} with
\eqref{eqn:qbow:final:meta} proves \eqref{eqn:qbow:aggregation} and completes the proof.
\subsection{Proof of Corollary~\ref{cor:bow:quadratic}}
For the competitive-ratio guarantee, applying
Theorem~\ref{thm:greedy:convex:quadratic} to $E^0$ gives
\begin{align*}
    \sum_{t=1}^T
    \left(
        f_t(\x_t)+\frac12\|\x_t-\x_{t-1}\|^2
    \right)     \overset{\eqref{eqn:qbow:aggregation}}{\leq} &
    \sum_{t=1}^T \left( f_t(\x_t^0) + \frac{1}{2}\|\x_t^0- \x_{t-1}^0\|^2 \right)+D^2
    \\
    \overset{\eqref{eqn:greedy:convex:quadratic}}{\leq}&
    \left(1+\frac{2}{\sqrt{\lambda}}\right)
    \sum_{t=1}^T
    \left(
        f_t(\u_t)+\frac12\|\u_t-\u_{t-1}\|^2
    \right)
    +D^2,
\end{align*}
for every comparator sequence $\u_0,\ldots,\u_T\in\X$
with $\u_0=\x_0$.

For the dynamic-regret guarantee, we use the following lemma.
\begin{lem}\label{lem:qbow:dynamic:expert}
Suppose Assumptions~\ref{ass:0} and~\ref{ass:2} hold.
For any initial point $\x_0\in\X$, let
$\x_1,\ldots,\x_T$ be generated by \eqref{eqn:2}
with $\gamma=1$.
Then, for every comparator sequence
$\u_0,\ldots,\u_T\in\X$,
\begin{equation}\label{eqn:qbow:dynamic:expert}
\begin{split}
&\sum_{t=1}^T
\left(f_t(\x_t)+\frac{1}{2}\|\x_t-\x_{t-1}\|^2\right)
-\sum_{t=1}^T
\left(f_t(\u_t)+\frac{1}{2}\|\u_t-\u_{t-1}\|^2\right)
\\
\leq&
\frac{1}{2}\|\u_0-\x_0\|^2
+D\sum_{t=1}^T\|\u_t-\u_{t-1}\|.
\end{split}
\end{equation}
\end{lem}
For any comparator sequence $\u_0,\ldots,\u_T\in\X$,
applying Lemma~\ref{lem:qbow:dynamic:expert} to $E^1$ gives
\begin{align*}
    &\sum_{t=1}^T
    \left(
        f_t(\x_t)+\frac12\|\x_t-\x_{t-1}\|^2
    \right)
    -\sum_{t=1}^T
    \left(
        f_t(\u_t)+\frac12\|\u_t-\u_{t-1}\|^2
    \right)
    \\
    \overset{\eqref{eqn:qbow:aggregation}}{\leq}&
    \sum_{t=1}^T  \left( f_t(\x_t^1) + \frac{1}{2}\|\x_t^1- \x_{t-1}^1\|^2 \right)
    -\sum_{t=1}^T
    \left(
        f_t(\u_t)+\frac12\|\u_t-\u_{t-1}\|^2
    \right)
    +8D^2\sqrt{T\log T}+3D^2
    \\
    \overset{\eqref{eqn:qbow:dynamic:expert}}{\leq}&
    \frac12\|\u_0-\x_0\|^2
    +DP_T
    +8D^2\sqrt{T\log T}+3D^2,
\end{align*}
which establishes \eqref{eqn:qbow:dynamic} and completes the proof.

\section{Conclusion and Future Work} \label{sec:conclusion}
This paper establishes competitive-ratio and dynamic-regret
guarantees for simple algorithms in smoothed online learning.
The naive algorithm achieves strong competitive guarantees
for polyhedral and quadratic growth hitting costs without
requiring convexity.
For polyhedral hitting costs, its guarantee is tight,
and the algorithm is optimal among deterministic online
algorithms for $\alpha\geq1$.
For convex quadratic growth hitting costs, the greedy
algorithm improves upon the naive guarantee when the
growth parameter is small.
For general convex hitting costs, SAder and its lookahead
counterpart achieve optimal dynamic regret in terms of
the horizon and the comparator's path-length.
Lookahead removes the bounded-gradient assumption without
improving the worst-case order of regret.

UDNP complements these results by combining competitive-ratio
and dynamic-regret guarantees under both the $\ell_2$-norm
and squared $\ell_2$-norm switching costs.
Using lookahead information, it handles potentially unbounded
expert-cost differences while controlling the additional
switching cost of aggregation.
Its guarantees apply to arbitrary pairs of experts,
allowing improvements in either expert to be incorporated
without redesigning the aggregation method.

Several directions deserve further investigation.
\begin{compactenum}
\item
A natural direction is to exploit polyhedrality or quadratic
growth of convex hitting costs to sharpen dynamic-regret
bounds under the $\ell_2$-norm switching cost and characterize
their dependence on the comparator's path-length and the
growth parameters.

\item
For $\alpha$-polyhedral hitting costs with $0<\alpha<1$,
the optimal deterministic competitive ratio remains open.
For quadratic growth hitting costs without convexity,
it remains open whether accounting for switching costs
can improve the naive guarantee.
Sharper lower bounds would help characterize the optimal
competitive ratios in these settings.
\end{compactenum}


\bibliography{ref}
\appendix
\section{Proof of Lemma~\ref{lem:meta}}
Based on the prediction rule of the meta-algorithm, we upper bound the switching cost when $t\geq 2$ as follows:
\begin{equation} \label{eqn:decompose:switching}
\begin{split}
& \|\x_t -\x_{t-1}\| =\left\| \sum_{\eta \in \H} w_t^\eta \x_t^\eta- \sum_{\eta \in \H} w_{t-1}^\eta \x_{t-1}^\eta \right\|  = \left\| \sum_{\eta \in \H} w_t^\eta (\x_t^\eta-\x)- \sum_{\eta \in \H} w_{t-1}^\eta (\x_{t-1}^\eta -\x) \right\|\\
\leq & \left\| \sum_{\eta \in \H} w_t^\eta (\x_t^\eta - \x)- \sum_{\eta \in \H} w_t^\eta (\x_{t-1}^\eta- \x) \right\| + \left\| \sum_{\eta \in \H} w_t^\eta (\x_{t-1}^\eta- \x) - \sum_{\eta \in \H} w_{t-1}^\eta (\x_{t-1}^\eta- \x) \right\| \\
=& \left\| \sum_{\eta \in \H} w_t^\eta (\x_t^\eta -   \x_{t-1}^\eta) \right\| + \left\| \sum_{\eta \in \H} (w_t^\eta -w_{t-1}^\eta)(\x_{t-1}^\eta- \x) \right\| \\
\leq &  \sum_{\eta \in \H} w_t^\eta \left\| \x_t^\eta -   \x_{t-1}^\eta \right\| + \sum_{\eta \in \H} |w_t^\eta -w_{t-1}^\eta|\left\| \x_{t-1}^\eta- \x  \right\| \\
\overset{(\ref{eqn:domain})}{\leq} &  \sum_{\eta \in \H} w_t^\eta \left\| \x_t^\eta -   \x_{t-1}^\eta \right\| + D \sum_{\eta \in \H} |w_t^\eta -w_{t-1}^\eta|= \sum_{\eta \in \H} w_t^\eta \left\| \x_t^\eta -   \x_{t-1}^\eta \right\| + D \|\w_t -\w_{t-1}\|_1
\end{split} \!\!\!\!\!\!\!\!\!\!\!
\end{equation}
where $\x$ is an arbitrary point in $\X$, and $\w_t=(w_t^{\eta})_{\eta \in \H} \in \R^N$. When $t=1$, from (\ref{eqn:condit:at:zero}), we have
\begin{equation} \label{eqn:decompose:switching:t1}
\|\x_1 -\x_{0}\| =  \left \| \sum_{\eta \in \H} w_1^\eta \big( \x_1^\eta -  \x_{0} \big)\right \| = \left \| \sum_{\eta \in \H} w_1^\eta \big( \x_1^\eta - \x_0^\eta \big) \right \|  \leq\sum_{\eta \in \H} w_1^\eta \left\| \x_1^\eta - \x_0^\eta \right\| .
\end{equation}

Then, the relative loss of the meta-algorithm w.r.t.~expert $E^\eta$ can be decomposed as
\begin{equation} \label{eqn:meta:loss}
\begin{split}
& \sum_{t=1}^T \big( s_t(\x_t) + \|\x_t -\x_{t-1}\|\big) - \sum_{t=1}^T \big(s_t(\x_t^\eta)  + \|\x_t^\eta- \x_{t-1}^\eta\| \big) \\
\overset{(\ref{eqn:surrogate}), (\ref{eqn:decompose:switching}),(\ref{eqn:decompose:switching:t1})}{\leq} &  \sum_{t=1}^T \left(\sum_{\eta \in \H} w_t^\eta  \big\| \x_t^\eta -   \x_{t-1}^\eta \big\|  - \big( \langle \nabla f_t(\x_t), \x_t^\eta -\x_t \rangle +  \|\x_t^\eta- \x_{t-1}^\eta\| \big) \right) \\
&+  D \sum_{t=2}^T  \|\w_t -\w_{t-1}\|_1 \\
\overset{(\ref{eqn:meta:loss:function})}{=} &  \underbrace{\sum_{t=1}^T \left(\sum_{\eta \in \H} w_t^\eta  \ell_t(\x_t^\eta)  - \ell_t(\x_t^\eta)  \right)}_{:=A}  +  D \sum_{t=2}^T  \|\w_t -\w_{t-1}\|_1.
\end{split}
\end{equation}

We proceed to bound $A$ and $\|\w_t -\w_{t-1}\|_1$ in (\ref{eqn:meta:loss}). Notice that $A$ is the regret of the meta-algorithm w.r.t.~expert $E^\eta$. From Assumptions~\ref{ass:2} and \ref{ass:1}, we have
\[
\left| \langle \nabla f_t(\x_t), \x_t^\eta -\x_t \rangle  \right| \leq \| \nabla f_t(\x_t)\| \|\x_t^\eta -\x_t\|  \overset{(\ref{eqn:grad}), (\ref{eqn:domain})}{\leq} GD.
\]
Thus, we have
\begin{equation} \label{eqn:bound:loss}
-GD \leq \ell_t(\x_t^\eta)  \leq (G+1)D, \ \forall \eta \in \H.
\end{equation}
According to the standard analysis of Hedge \citep[Lemma 1]{Adaptive:Dynamic:Regret:NIPS} and (\ref{eqn:bound:loss}), we have
\begin{equation} \label{eqn:bound:A}
\sum_{t=1}^T \left(\sum_{\eta \in \H} w_t^\eta  \ell_t(\x_t^\eta)  - \ell_t(\x_t^\eta)  \right) \leq \frac{1}{\beta} \ln \frac{1}{w_1^\eta}  + \frac{\beta T (2G+1)^2 D^2}{8}.
\end{equation}

Next, we bound $\|\w_t -\w_{t-1}\|_1$, which measures the stability of the meta-algorithm, i.e., the change of coefficients between successive rounds.  Because the Hedge algorithm is translation invariant, we can subtract $D/2$ from $\ell_t(\x_t^\eta)$ such that
\begin{equation} \label{eqn:bound:loss2}
\left| \ell_t(\x_t^\eta) - D/2 \right| \leq (G+1/2)D, \ \forall \eta \in \H.
\end{equation}
It is well-known that Hedge can be treated as a special case of ``Follow-the-Regularized-Leader'' with  entropic regularization \citep{Online:suvery}
\[
R(\w)=\sum_i w_i \log w_i
\]
over the probability simplex, and $R(\cdot)$ is $1$-strongly convex  w.r.t.~the $\ell_1$-norm. In other words, we have
 \[
 \w_{t+1}= \argmin_{\w \in \Delta} \left \langle - \frac{1}{\beta} \log (\w_1) + \sum_{i=1}^t \g_i , \w \right \rangle + \frac{1}{\beta} R(\w), \ \forall t \geq 1
 \]
 where $\Delta \subseteq \R^{N}$ is the probability simplex, and $\g_i=[\ell_i(\x_i^\eta)- D/2]_{\eta \in \H} \in \R^N$.  From the stability property of Follow-the-Regularized-Leader \citep[Lemma 2]{DV:Distributed}, we have
\[
\|\w_t -\w_{t-1}\|_1 \leq \beta \|\g_{t-1}\|_{\infty} \overset{(\ref{eqn:bound:loss2})}{\leq} \beta (G+1/2)D, \ \forall t \geq 2.
\]
Then
\begin{equation}\label{eqn:bound:sum:b}
 \sum_{t=2}^T \|\w_t -\w_{t-1}\|_1 \leq  \frac{ \beta (T-1)(2G +1)D}{2}.
\end{equation}

Substituting  (\ref{eqn:bound:A}) and (\ref{eqn:bound:sum:b}) into (\ref{eqn:meta:loss}), we have
\[
\begin{split}
& \sum_{t=1}^T \big( s_t(\x_t) + \|\x_t -\x_{t-1}\|\big) - \sum_{t=1}^T \big(s_t(\x_t^\eta)  + \|\x_t^\eta- \x_{t-1}^\eta\| \big)\\
\leq  &  \frac{1}{\beta} \ln \frac{1}{w_1^\eta} + \frac{\beta T (2G+1)^2 D^2}{8} + \frac{ \beta (T-1)(2G +1)D^2}{2}   \leq \frac{1}{\beta} \ln \frac{1}{w_1^\eta} + \frac{5 \beta T (2G+1)^2 D^2}{8}.
\end{split}
\]
We complete the proof by setting $\beta= \frac{2}{(2G+1) D} \sqrt{\frac{2 }{5 T}}$.
\section{Proof of Lemma~\ref{lem:expert}}
First, we bound the dynamic regret of the expert-algorithm. Define
\[
\xb_{t+1}^\eta=\x_t^\eta - \eta \nabla f_t(\x_t).
\]
Following the analysis of Ader \citep[Theorems 1 and 6]{Adaptive:Dynamic:Regret:NIPS}, we have
\[
\begin{split}
& s_t(\x_t^\eta) - s_t(\u_t)\overset{(\ref{eqn:surrogate})}{=} \langle \nabla f_t(\x_t), \x_t^\eta - \u_t\rangle = \frac{1}{\eta} \langle \x_t^\eta  - \xb_{t+1}^\eta, \x_t^\eta - \u_t\rangle \\
= & \frac{1}{2 \eta} \left( \|\x_t^\eta-\u_t\|^2 - \|\xb_{t+1}^\eta-\u_t\|^2 + \|\x_t^\eta  - \xb_{t+1}^\eta\|^2 \right)\\
 =& \frac{1}{2 \eta} \left( \|\x_t^\eta-\u_t\|^2 - \|\xb_{t+1}^\eta-\u_t\|^2 \right) + \frac{\eta}{2 } \|\nabla f_t(\x_t)\|^2  \\
\overset{(\ref{eqn:grad})}{\leq} & \frac{1}{2 \eta} \left( \|\x_t^\eta-\u_t\|^2 - \|\xb_{t+1}^\eta-\u_t\|^2 \right) + \frac{\eta}{2 } G^2  \\
\leq & \frac{1}{2 \eta} \left( \|\x_t^\eta-\u_t\|^2 - \|\x_{t+1}^\eta-\u_t\|^2 \right) + \frac{\eta}{2 } G^2 \\
= & \frac{1}{2 \eta} \left( \|\x_t^\eta-\u_t\|^2 - \|\x_{t+1}^\eta-\u_{t+1}\|^2 + \|\x_{t+1}^\eta-\u_{t+1}\|^2 -\|\x_{t+1}^\eta-\u_t\|^2 \right) + \frac{\eta}{2 } G^2 \\
= & \frac{1}{2 \eta} \left( \|\x_t^\eta-\u_t\|^2 - \|\x_{t+1}^\eta-\u_{t+1}\|^2 + (\x_{t+1}^\eta-\u_{t+1} +\x_{t+1}^\eta-\u_t )^\top (\u_t-\u_{t+1})  \right) + \frac{\eta}{2 } G^2 \\
\leq  & \frac{1}{2 \eta} \left( \|\x_t^\eta-\u_t\|^2 - \|\x_{t+1}^\eta-\u_{t+1}\|^2 + \big(\|\x_{t+1}^\eta-\u_{t+1}\| + \|\x_{t+1}^\eta-\u_t \|\big) \|\u_t-\u_{t+1}\| \right) + \frac{\eta}{2 } G^2 \\
\overset{(\ref{eqn:domain})}{\leq}  & \frac{1}{2 \eta} \left( \|\x_t^\eta-\u_t\|^2 - \|\x_{t+1}^\eta-\u_{t+1}\|^2 \right) + \frac{D}{ \eta} \|\u_t-\u_{t+1}\| + \frac{\eta}{2 } G^2 .
\end{split}
\]
Summing the above inequality over all iterations, we have
\begin{equation} \label{eqn:expert:1}
\begin{split}
\sum_{t=1}^T \left( s_t(\x_t^\eta) - s_t(\u_t) \right) \leq & \frac{1}{2 \eta}   \|\x_1^\eta-\u_1\|^2  + \frac{D}{ \eta} \sum_{t=1}^T \|\u_{t+1} - \u_t\| + \frac{\eta T}{2 } G^2 \\
\overset{(\ref{eqn:domain})}{\leq} &  \frac{1}{2 \eta} D^2    + \frac{D}{ \eta} \sum_{t=1}^T \|\u_{t+1} - \u_t\| + \frac{\eta T}{2 } G^2.
\end{split}
\end{equation}
Since  (\ref{eqn:expert:1})  holds when $\u_{T+1} =\u_T$, we have
\begin{equation} \label{eqn:expert:2}
\begin{split}
\sum_{t=1}^T \left( s_t(\x_t^\eta) - s_t(\u_t) \right) \leq  \frac{1}{2 \eta} D^2    + \frac{D}{ \eta} \sum_{t=1}^T \|\u_t - \u_{t-1}\| + \frac{\eta T}{2 } G^2.
\end{split}
\end{equation}

Next, we bound the switching cost of the expert-algorithm. Since $\x_1^\eta = \x_{0}^\eta$, we have 
\begin{equation} \label{eqn:expert:3}
\begin{split}
& \sum_{t=1}^T \|\x_t^\eta- \x_{t-1}^\eta\|  = \sum_{t=1}^{T-1} \|\x_{t+1}^\eta-\x_{t}^\eta\| \leq   \sum_{t=1}^{T-1} \|\xb_{t+1}^\eta-\x_{t}^\eta\| =  \sum_{t=1}^{T-1} \|\eta \nabla f_t(\x_t)\| \overset{(\ref{eqn:grad})}{\leq} \eta (T-1) G.
\end{split}
\end{equation}
We complete the proof by combining (\ref{eqn:expert:2}) with (\ref{eqn:expert:3}).
\section{Proof of Lemma~\ref{lem:lookhead:meta}}
Similar to the proof of Lemma~\ref{lem:meta}, we  first relate the one-round cost of the meta-algorithm to the weighted
one-round costs of the experts. By convexity of $f_t(\cdot)$,
\begin{equation}\label{eqn:lookahead:hitting:interpolation}
f_t(\x_t) =f_t\left(\sum_{\eta\in\H}w_t^\eta\x_t^\eta\right) \leq \sum_{\eta\in\H} w_t^\eta f_t(\x_t^\eta).
\end{equation}
Then, we have
\begin{equation} \label{eqn:meta:lookahead:loss}
\begin{split}
& \sum_{t=1}^T \big( f_t(\x_t) + \|\x_t -\x_{t-1}\|\big) - \sum_{t=1}^T \big(f_t(\x_t^\eta)  + \|\x_t^\eta- \x_{t-1}^\eta\| \big) \\
\overset{(\ref{eqn:lookahead:hitting:interpolation}), (\ref{eqn:decompose:switching}),(\ref{eqn:decompose:switching:t1})}{\leq} &  \sum_{t=1}^T \left(\sum_{\eta \in \H} w_t^\eta \big( f_t(\x_t^\eta) +\| \x_t^\eta -   \x_{t-1}^\eta \| \big) - \big( f_t(\x_t^\eta)  +  \|\x_t^\eta- \x_{t-1}^\eta\| \big) \right) \\
&+  D \sum_{t=2}^T  \|\w_t -\w_{t-1}\|_1 \\
\overset{(\ref{eqn:lookahead:loss})}{=} &  \underbrace{\sum_{t=1}^T \left(\sum_{\eta \in \H} w_t^\eta  c_t(\x_t^\eta)  - c_t(\x_t^\eta)  \right)}_{:=A}  +  D \sum_{t=2}^T  \|\w_t -\w_{t-1}\|_1.
\end{split}
\end{equation}

To bound $A$, we need to analyze the behavior of the lookahead Hedge. To this end, we prove the following lemma. 
\begin{lem}\label{lem:lookhead:hedge} The meta-algorithm in Algorithm~\ref{alg:3} satisfies
\begin{equation} \label{eqn:lookhead:hedge}
\sum_{t=1}^T \left(\sum_{\eta \in \H} w_t^\eta  c_t(\x_t^\eta)  - c_t(\x_t^\eta)  \right)  \leq  \frac{1}{\beta} \ln \frac{1}{w_0^\eta} - \frac{1}{2\beta} \sum_{t=1}^T \|\w_t-\w_{t-1}\|_1^2
\end{equation}
for any $\eta \in \H$.
\end{lem}

Substituting (\ref{eqn:lookhead:hedge}) into (\ref{eqn:meta:lookahead:loss}), we have
\begin{equation} \label{eqn:meta:loss:new}
\begin{split}
& \sum_{t=1}^T \big( f_t(\x_t) + \|\x_t -\x_{t-1}\|\big) - \sum_{t=1}^T \big(f_t(\x_t^\eta)  + \|\x_t^\eta- \x_{t-1}^\eta\| \big) \\
\leq & \frac{1}{\beta} \ln \frac{1}{w_0^\eta} - \frac{1}{2\beta} \sum_{t=1}^T \|\w_t-\w_{t-1}\|_1^2  +  D \sum_{t=2}^T  \|\w_t -\w_{t-1}\|_1 \\
\leq & \frac{1}{\beta} \ln \frac{1}{w_0^\eta} - \frac{1}{2\beta} \sum_{t=1}^T \|\w_t-\w_{t-1}\|_1^2  +   \sum_{t=2}^T \left( \frac{1}{2\beta} \|\w_t -\w_{t-1}\|_1^2 + \frac{\beta D^2}{2} \right) \\
\leq &  \frac{1}{\beta} \ln \frac{1}{w_0^\eta} + \frac{\beta T D^2}{2} =D \sqrt{ \frac{T}{2}}  \left( \ln \frac{1}{w_0^\eta} +1 \right)
\end{split}
\end{equation}
where we set $\beta= \frac{1}{D}  \sqrt{\frac{2}{T}}$.

\section{Proof of Lemma~\ref{lem:lookhead:hedge}}
To simplify the notation, we define
\[
W_0= \sum_{\eta \in \H} w_0^\eta =1, \ L_t^\eta= \sum_{i=1}^t c_i(\x_i^\eta), \textrm{ and } W_t= \sum_{\eta \in \H} w_0^\eta e^{- \beta L_t^\eta}, \ \forall t \geq 1.
\]
From the updating rule in (\ref{eqn:lookahead:hedge:pro}), it is easy to verify that
\begin{equation} \label{eqn:wt:eta}
w_t^\eta =  \frac{w_0^\eta  e^{-\beta L_{t}^\eta}}{W_t}, \ \forall t \geq 1.
\end{equation}

First, we have
\begin{equation} \label{eqn:wt:w0}
\begin{split}
\ln W_T = \ln \left( \sum_{\eta \in \H} w_0^\eta e^{- \beta L_T^\eta}\right) \geq \ln\left( \max_{\eta \in \H}  w_0^\eta e^{- \beta L_T^\eta} \right) =- \beta \min_{\eta \in \H} \left(L_T^\eta + \frac{1}{\beta} \ln \frac{1}{w_0^\eta} \right) .
\end{split}
\end{equation}

Next, we bound the related quantity $\ln(W_t/W_{t-1})$ as follows.  For any $\eta \in \H$, we have
\begin{equation} \label{eqn:wt:wt1}
\ln \left(\frac{W_t}{W_{t-1}} \right)\overset{(\ref{eqn:wt:eta})}{=} \ln \left(\frac{w_0^\eta  e^{-\beta L_{t}^\eta}}{w_t^\eta}  \frac{w_{t-1}^\eta}{w_0^\eta  e^{-\beta L_{t-1}^\eta}} \right)  = \ln\left(\frac{w_{t-1}^\eta}{w_t^\eta} \right) - \beta c_t(\x_t^\eta).
\end{equation}
Then, we have
\begin{equation} \label{eqn:w1}
\begin{split}
&\ln \left(\frac{W_t}{W_{t-1}} \right) = \ln \left(\frac{W_t}{W_{t-1}} \right) \sum_{\eta \in \H} w_t^\eta =  \sum_{\eta \in \H} w_t^\eta \ln \left(\frac{W_t}{W_{t-1}} \right) \\
\overset{(\ref{eqn:wt:wt1})}{=} &\sum_{\eta \in \H} w_t^\eta \ln\left(\frac{w_{t-1}^\eta}{w_t^\eta} \right) - \beta \sum_{\eta \in \H} w_t^\eta c_t(\x_t^\eta) \leq - \frac{1}{2} \|\w_t-\w_{t-1}\|_1^2 - \beta \sum_{\eta \in \H} w_t^\eta c_t(\x_t^\eta)
\end{split}
\end{equation}
where the last inequality is due to Pinsker's inequality \citep[Lemma 11.6.1]{Elements:IT}. Thus
\begin{equation} \label{eqn:WT}
\ln W_T= \ln W_0 +\sum_{t=1}^T\ln \left(\frac{W_t}{W_{t-1}}\right) \overset{(\ref{eqn:w1})}{\leq}\sum_{t=1}^T  \left(- \frac{1}{2} \|\w_t-\w_{t-1}\|_1^2 - \beta \sum_{\eta \in \H} w_t^\eta c_t(\x_t^\eta) \right).
\end{equation}

Combining (\ref{eqn:wt:w0}) with (\ref{eqn:WT}), we obtain
\[
- \beta \min_{\eta \in \H} \left(L_T^\eta + \frac{1}{\beta} \ln \frac{1}{w_0^\eta} \right) \leq \sum_{t=1}^T  \left(- \frac{1}{2} \|\w_t-\w_{t-1}\|_1^2 - \beta \sum_{\eta \in \H} w_t^\eta c_t(\x_t^\eta) \right)
\]
Rearranging the above inequality and noting that the minimum is no larger than the same expression evaluated at any fixed $\eta\in\H$, we obtain (\ref{eqn:lookhead:hedge}) for every $\eta\in\H$.
\section{Proof of Lemma~\ref{lem:lookhead:expert}}
The analysis is similar to that of Theorem 10 of \citet{SOCO:OBD}, which relies on a strong condition
\[
\x_t^\eta=\x_{t-1}^\eta - \eta \nabla f_t(\x_t^\eta).
\]
This identity is the first-order optimality condition for  the unconstrained problem in \eqref{eqn:lookhead:expert};   it need not hold at a constrained minimizer.  In contrast, our
analysis does not require differentiability. Since $f_t(\cdot)$ is convex, the objective in (\ref{eqn:lookhead:expert}) is $1/\eta$-strongly convex and is
minimized at $\x_t^\eta$. Applying the quadratic growth property of strongly convex functions \citep{COLT:Hazan:2011}, as in
(\ref{eqn:opt:convex:quadratic}), we obtain
\[
f_t(\u_t)
+\frac{1}{2\eta}\|\x_{t-1}^\eta-\u_t\|^2 \geq f_t(\x_t^\eta) +\frac{1}{2\eta}\|\x_t^\eta-\x_{t-1}^\eta\|^2 +\frac{1}{2\eta}\|\x_t^\eta-\u_t\|^2
\]
for any $\u_t\in\X$. Rearranging this inequality gives 
\[
\begin{split}
 & f_t(\x_t^\eta)-f_t(\u_t) \\
\leq &  \frac{1}{2 \eta} \left( \|\x_{t-1}^\eta-\u_t \|^2 - \|\x_{t}^\eta-\u_t\|^2 - \|\x_t^\eta-\x_{t-1}^\eta\|^2\right) \\
= &\frac{1}{2 \eta} \left( \|\x_{t-1}^\eta-\u_{t-1} \|^2 - \|\x_{t}^\eta-\u_t\|^2 +  \|\x_{t-1}^\eta-\u_{t} \|^2 - \|\x_{t-1}^\eta-\u_{t-1} \|^2  - \|\x_t^\eta-\x_{t-1}^\eta\|^2\right) \\
=& \frac{1}{2 \eta} \left( \|\x_{t-1}^\eta-\u_{t-1} \|^2 - \|\x_{t}^\eta-\u_t\|^2 + \langle \x_{t-1}^\eta-\u_t + \x_{t-1}^\eta-\u_{t-1},\u_{t-1}- \u_t \rangle- \|\x_t^\eta-\x_{t-1}^\eta\|^2\right)  \\
\leq & \frac{1}{2 \eta} \left( \|\x_{t-1}^\eta-\u_{t-1} \|^2 - \|\x_{t}^\eta-\u_t\|^2 + \big(\| \x_{t-1}^\eta-\u_t \| +\| \x_{t-1}^\eta-\u_{t-1}\| \big) \|\u_t - \u_{t-1}\|\right)\\
& -  \frac{1}{2 \eta} \|\x_t^\eta-\x_{t-1}^\eta\|^2 \\
\overset{(\ref{eqn:domain})}{\leq}  &  \frac{1}{2 \eta} \left( \|\x_{t-1}^\eta-\u_{t-1} \|^2 - \|\x_{t}^\eta-\u_t\|^2 \right) + \frac{D}{\eta}\|\u_t - \u_{t-1}\| -  \frac{1}{2 \eta} \|\x_t^\eta-\x_{t-1}^\eta\|^2 .
\end{split}
\]

Summing the above inequality over all iterations, we have
\begin{equation} \label{eqn:greedy:2}
\begin{split}
\sum_{t=1}^T \left( f_t(\x_t^\eta) - f_t(\u_t) \right) \leq & \frac{1}{2 \eta}   \|\x_0^\eta-\u_0\|^2  + \frac{D}{\eta} \sum_{t=1}^T \|\u_t - \u_{t-1}\| -  \frac{1}{2 \eta} \sum_{t=1}^T \|\x_t^\eta-\x_{t-1}^\eta\|^2 \\
\overset{(\ref{eqn:domain})}{\leq} & \frac{1}{2 \eta}   D^2 + \frac{D}{\eta} \sum_{t=1}^T \|\u_t - \u_{t-1}\| -  \frac{1}{2 \eta} \sum_{t=1}^T \|\x_t^\eta-\x_{t-1}^\eta\|^2.
\end{split}
\end{equation}
We next bound the sum of ordinary dynamic regret and the learner's switching cost:
\[
\begin{split}
&\sum_{t=1}^T \left( f_t(\x_t^\eta) +\|\x_t^\eta-\x_{t-1}^\eta\|- f_t(\u_t) \right) \\
\overset{(\ref{eqn:greedy:2})}{\leq} & \frac{1}{2 \eta}   D^2  + \frac{D}{\eta} \sum_{t=1}^T \|\u_t - \u_{t-1}\|  -  \frac{1}{2 \eta} \sum_{t=1}^T \|\x_t^\eta-\x_{t-1}^\eta\|^2 +\sum_{t=1}^T  \|\x_t^\eta-\x_{t-1}^\eta\|\\
\leq & \frac{1}{2 \eta}   D^2  + \frac{D}{\eta} \sum_{t=1}^T \|\u_t - \u_{t-1}\| -  \frac{1}{2 \eta} \sum_{t=1}^T \|\x_t^\eta-\x_{t-1}^\eta\|^2 +\sum_{t=1}^T  \left( \frac{1}{2 \eta} \|\x_t^\eta-\x_{t-1}^\eta\|^2 + \frac{\eta}{2} \right) \\
= &\frac{1}{2 \eta}   D^2  + \frac{D}{\eta} \sum_{t=1}^T \|\u_t - \u_{t-1}\|  + \frac{\eta T}{2}.
\end{split}
\]

\section{Proof of Lemma~\ref{lem:bow:interpolation}} \label{app:proof:bow:interpolation}
The proof follows the standard interpolation argument in \citet[Lemma~1]{Smooth:DNP:NeurIPS}. For completeness, we provide the
details below.

By convexity of $f_t(\cdot)$, we have
\begin{equation}\label{eqn:bow:hitting:interpolation}
	f_t(\x_t) = f_t\big((1-p_t)\x_t^0+p_t\x_t^1\big) \le (1-p_t)f_t(\x_t^0)+p_tf_t(\x_t^1)
. \end{equation}
Moreover,
\[
	\x_t-\x_{t-1} = (1-p_t)(\x_t^0-\x_{t-1}^0) + p_t(\x_t^1-\x_{t-1}^1) + (p_t-p_{t-1})(\x_{t-1}^1-\x_{t-1}^0)
. \]
The triangle inequality and Assumption~\ref{ass:2} therefore imply
\begin{equation}\label{eqn:bow:movement:interpolation}
	\|\x_t-\x_{t-1}\|
	\le
	(1-p_t)\|\x_t^0-\x_{t-1}^0\| +p_t\|\x_t^1-\x_{t-1}^1\| +D|p_t-p_{t-1}|
. \end{equation}
Combining \eqref{eqn:bow:hitting:interpolation} with \eqref{eqn:bow:movement:interpolation} and using the definition of $C_t^i$ in \eqref{eqn:bow:expert:cost}, we obtain
\begin{equation}\label{eqn:bow:cost:interpolation}
	f_t(\x_t)+\|\x_t-\x_{t-1}\| \le (1-p_t) C_t^0+p_tC_t^1+D|p_t-p_{t-1}|
. \end{equation}

Summing over $t=1,\ldots,T$, we obtain
\[
\sum_{t=1}^{T} \big(f_t(\x_t)+\|\x_t-\x_{t-1}\|
\big) \leq \sum_{t=1}^{T}
\left(
(1-p_t)C_t^0+p_tC_t^1
\right)+
\sum_{t=1}^{T}
D|p_t-p_{t-1}|.
\]
Using
\[
(1-p_t)C_t^0+p_tC_t^1
=
C_t^0-p_t\Delta_t
=
C_t^1+(1-p_t)\Delta_t,
\]
we obtain the two claimed inequalities
\eqref{eq:bow:interpolation:E0} and
\eqref{eq:bow:interpolation:E1}.

\section{Proof of Lemma~\ref{lem:qbow:interpolation}} \label{app:proof:qbow:interpolation} 
By convexity of $f_t(\cdot)$, the hitting-cost bound \eqref{eqn:bow:hitting:interpolation} holds.
To bound the switching cost, we consider two cases.

\textbf{Case 1: $p_t\geq p_{t-1}$.}
We write
\[
\x_t-\x_{t-1}=(1-p_t)(\x_t^0-\x_{t-1}^0)+p_{t-1}(\x_t^1-\x_{t-1}^1)+(p_t-p_{t-1})(\x_t^1-\x_{t-1}^0).
\]
The three coefficients are nonnegative and sum to one.
By convexity of the squared norm and
Assumption~\ref{ass:2}, we obtain
\[
\begin{aligned}
\frac12\|\x_t-\x_{t-1}\|^2
&\leq
\frac{1-p_t}{2}\|\x_t^0-\x_{t-1}^0\|^2
+\frac{p_{t-1}}{2}\|\x_t^1-\x_{t-1}^1\|^2
+\frac{D^2}{2}(p_t-p_{t-1})\\
&\leq
\frac{1-p_t}{2}\|\x_t^0-\x_{t-1}^0\|^2
+\frac{p_t}{2}\|\x_t^1-\x_{t-1}^1\|^2
+\frac{D^2}{2}(p_t-p_{t-1}),
\end{aligned}
\]
where the second inequality uses $p_{t-1}\leq p_t$.

\textbf{Case 2: $p_t<p_{t-1}$.}
In this case, we write
\[
\x_t-\x_{t-1}=(1-p_{t-1})(\x_t^0-\x_{t-1}^0)+p_t(\x_t^1-\x_{t-1}^1)+(p_{t-1}-p_t)(\x_t^0-\x_{t-1}^1).
\]
Applying the same argument and using
$1-p_{t-1}\leq 1-p_t$ gives
\[
\frac12\|\x_t-\x_{t-1}\|^2
\leq
\frac{1-p_t}{2}\|\x_t^0-\x_{t-1}^0\|^2
+\frac{p_t}{2}\|\x_t^1-\x_{t-1}^1\|^2
+\frac{D^2}{2}(p_{t-1}-p_t).
\]

Thus, in either case, combining the switching-cost bound
with \eqref{eqn:bow:hitting:interpolation}
and using the definition of $C_t^i$ in
\eqref{eq:bow:squared_l2:expert:cost}, we obtain
\begin{equation}\label{eqn:qbow:interpolation}
f_t(\x_t)+\frac12\|\x_t-\x_{t-1}\|^2
\leq
(1-p_t)C_t^0+p_tC_t^1
+\frac{D^2}{2}|p_t-p_{t-1}|.
\end{equation}

Summing over $t=1,\ldots,T$, we obtain
\[
\sum_{t=1}^{T}
\left(f_t(\x_t)+\frac12\|\x_t-\x_{t-1}\|^2\right)
\leq
\sum_{t=1}^{T}\left((1-p_t)C_t^0+p_tC_t^1\right)
+\frac{D^2}{2}\sum_{t=1}^{T}|p_t-p_{t-1}|.
\]
Using
\[
(1-p_t)C_t^0+p_tC_t^1
=C_t^0-p_t\Delta_t
=C_t^1+(1-p_t)\Delta_t,
\]
we obtain the two claimed inequalities
\eqref{eq:qbow:interpolation:E0} and
\eqref{eq:qbow:interpolation:E1}.

\section{Proof of Lemma~\ref{lem:bow:transfer}}\label{app:bow:transfer}
We prove the two inequalities in \eqref{eqn:bow:transfer} in turn.

\paragraph{First inequality.}
For every $t$ and $i\in\{0,1\}$, the weight-selection rule
in Lemma~\ref{lem:bow:transfer} implies
\[
	(i-p_t)(\Delta_t-\dtb_t) \le 0
. \]
Indeed, when $\Delta_t\geq K$, we have $p_t=1$ and $\Delta_t-\dtb_t\geq0$; when $\Delta_t\leq-K$, we have $p_t=0$ and $\Delta_t-\dtb_t\le0$.
Otherwise, $\Delta_t=\dtb_t$.
Since clipping does not change the switching term, it follows that
\begin{equation}\label{eqn:bow:unclip}
	\Rt_i(\p;\Delta;K)-\Rt_i(\p;\dtb;K)
	=
	\sum_{t=1}^T(i-p_t)(\Delta_t-\dtb_t)
	\le
	0
. \end{equation}

\paragraph{Second inequality.}
Let $e_t=p_t-q_t$ for $t=0,\ldots,T$, so that $e_0=0$. The clipping and weight-selection rules in
Lemma~\ref{lem:bow:transfer} give
\begin{equation}\label{eqn:bow:gap:gain}
	e_t\dtb_t = K|e_t|
. \end{equation}
To see this, if $\Delta_t\ge K$, then $e_t=1-q_t\ge 0$ and $\dtb_t=K$.
If $\Delta_t\le -K$, then $e_t=-q_t\le 0$ and $\dtb_t=-K$.
In all other cases, $e_t=0$.

The triangle inequality and $e_0=0$ imply
\begin{equation}	\label{eqn:bow:variation}
\begin{split}
\sum_{t=1}^T \big( |p_t-p_{t-1}|-|q_t-q_{t-1}| \big)
	&=
	\sum_{t=1}^T \big( |(q_t-q_{t-1})+(e_t-e_{t-1})|-|q_t-q_{t-1}| \big)
	\\
	&\le
	\sum_{t=1}^T \big( |q_t-q_{t-1}|+|e_t-e_{t-1}|-|q_t-q_{t-1}| \big)
	\\
	&= \sum_{t=1}^T|e_t-e_{t-1}|\le \sum_{t=1}^T\left(|e_t|+|e_{t-1}|\right)\le 2\sum_{t=1}^T|e_t|. 
  \end{split}    
\end{equation}
Consequently, for either $i\in\{0,1\}$,
\begin{equation} 	\label{eqn:bow:endpoint:transfer}
\begin{split}
 &	\Rt_i(\p;\dtb;K)-\Rt_i(\q;\dtb;K) \\
=& 	\frac{K}{2} \sum_{t=1}^T \left( |p_t-p_{t-1}|-|q_t-q_{t-1}| \right)
	- \sum_{t=1}^T e_t\dtb_t  	\\
    \le  &	K\sum_{t=1}^T|e_t| - 	K\sum_{t=1}^T|e_t|=0,
\end{split}
\end{equation}
where the inequality follows from \eqref{eqn:bow:gap:gain} and \eqref{eqn:bow:variation}.
Combining \eqref{eqn:bow:unclip} and \eqref{eqn:bow:endpoint:transfer} proves the lemma.

\section{Proof of Lemma~\ref{lem:bow:clip_bound}} \label{app:proof:bow:clip_bound}
For the analysis, after the final update at round $T$, define $q_{T+1}:=g(\theta_{T+1})$. For $Z=1/T$, the assumption $T\geq\max\{8e,16\log T\}$ ensures that $Z\leq1/e$ and $T\geq\max\{8e,16\log(1/Z)\}$, as required by the theoretical guarantee of DNP, i.e., Theorem~5 of \citet{Smooth:DNP:NeurIPS}. Applying this result (with the notation therein and $\mu=1, r=1, s=n=\tau=T$, and $x_1=0$) gives
\begin{equation}\label{eqn:bow:dnp}
\sum_{t=1}^T\big(q_tb_t-|q_t-q_{t+1}|\big) \geq \max\left\{0,\sum_{t=1}^Tb_t- 2U_T-2 \right\}-1
\end{equation}
where
\begin{equation}\label{eqn:bow:dnp:ut}
U_T=\tg^{-1}(1) \leq  4\sqrt{T\log T}.
\end{equation}
Substituting $b_t=\dtb_t/K$ into \eqref{eqn:bow:dnp} and multiplying both sides by $K$, we obtain
\begin{equation}\label{eqn:bow:dnp:scaled}
\sum_{t=1}^T\big(q_t \dtb_t-K |q_t-q_{t+1}|\big) \ge \max\left\{0,\sum_{t=1}^T\dtb_t-2KU_T-2K \right\}-K
. \end{equation}
Since $q_0=q_1=g(0)=0$, we have
\begin{equation}\label{eqn:bow:dnp:switching}
\frac{K}{2}\sum_{t=1}^T|q_t-q_{t-1}|
=
\frac{K}{2}\sum_{t=1}^{T-1}|q_{t+1}-q_t|
\le
K\sum_{t=1}^T|q_{t+1}-q_t|.
\end{equation}
Combining \eqref{eqn:bow:dnp:scaled}
with \eqref{eqn:bow:dnp:switching}
gives the following bounds on the two reference residuals:
\begin{align*}
\Rt_0(\q;\dtb;K)
& \overset{\eqref{eqn:bow:dnp:switching}}{\le}
\sum_{t=1}^T\big(K|q_t-q_{t+1}|-q_t\dtb_t\big)
\overset{(\ref{eqn:bow:dnp:scaled})}{\leq}
K,
\\
\Rt_1(\q;\dtb;K)
&\overset{\eqref{eqn:bow:dnp:switching}}{\le}
\sum_{t=1}^T\left(K|q_t-q_{t+1}|+(1-q_t)\dtb_t\right)
\overset{\eqref{eqn:bow:dnp:ut},\eqref{eqn:bow:dnp:scaled}}{\le}
8K\sqrt{T\log T}+3K.
\end{align*}

\section{Proof of Lemma~\ref{lem:qbow:dynamic:expert}}
\label{app:lem:qbow:dynamic:expert}
The proof of Theorem~6 of \citet{NIPS2019_8463}
uses a vanishing gradient condition for the regularized
objective, which need not hold at a constrained minimizer.
We therefore provide a direct proof that accommodates
nonsmooth convex hitting costs and constrained minimizers.

Since $f_t(\cdot)$ is convex, the objective in
\eqref{eqn:2} with $\gamma=1$ is $1$-strongly convex
and is minimized at $\x_t$.
Applying the quadratic growth property of strongly convex
functions, as in \eqref{eqn:opt:convex:quadratic}, we obtain
\begin{equation}\label{eqn:qbow:dynamic:expert:one}
    f_t(\x_t)
    +\frac12\|\x_t-\x_{t-1}\|^2
    +\frac12\|\u_t-\x_t\|^2
    \leq
    f_t(\u_t)
    +\frac12\|\u_t-\x_{t-1}\|^2
\end{equation}
for every $\u_t\in\X$.

Using
$\u_t-\x_{t-1}
=(\u_{t-1}-\x_{t-1})+(\u_t-\u_{t-1})$,
we obtain from \eqref{eqn:qbow:dynamic:expert:one} that
\begin{align*}
    &f_t(\x_t)
    +\frac12\|\x_t-\x_{t-1}\|^2
    -f_t(\u_t)
    -\frac12\|\u_t-\u_{t-1}\|^2
    \\
    \leq&
    \frac12\|\u_t-\x_{t-1}\|^2
    -\frac12\|\u_t-\x_t\|^2
    -\frac12\|\u_t-\u_{t-1}\|^2
    \\
    =&
    \frac12\|\u_{t-1}-\x_{t-1}\|^2
    -\frac12\|\u_t-\x_t\|^2
    +\left\langle
        \u_{t-1}-\x_{t-1},\u_t-\u_{t-1}
    \right\rangle
    \\
    \leq&
    \frac12\|\u_{t-1}-\x_{t-1}\|^2
    -\frac12\|\u_t-\x_t\|^2
    +D\|\u_t-\u_{t-1}\|,
\end{align*}
where the last inequality follows from the
Cauchy--Schwarz inequality and Assumption~\ref{ass:2}.
Summing over $t=1,\ldots,T$ and telescoping gives
\begin{align*}
    &\sum_{t=1}^T
    \left(
        f_t(\x_t)+\frac12\|\x_t-\x_{t-1}\|^2
    \right)
    -\sum_{t=1}^T
    \left(
        f_t(\u_t)+\frac12\|\u_t-\u_{t-1}\|^2
    \right)
    \\
    \leq &
    \frac12\|\u_0-\x_0\|^2
    -\frac12\|\u_T-\x_T\|^2
    +D\sum_{t=1}^T\|\u_t-\u_{t-1}\|
    \\
    \leq&
    \frac12\|\u_0-\x_0\|^2
    +D\sum_{t=1}^T\|\u_t-\u_{t-1}\|,
\end{align*}
which proves the lemma.
\end{document}